\documentclass[journal]{IEEEtran}

\usepackage{hyperref}
\usepackage{url}
\usepackage{amsfonts}
\usepackage{amsmath}
\usepackage{varioref}
\usepackage{amssymb}
\usepackage{bm}
\usepackage{color}
\usepackage{subfig}
\usepackage{graphicx}
\usepackage{booktabs}
\usepackage{multirow}
\usepackage[]{algorithm2e}
\DeclareMathOperator*{\argmin}{arg\,min}
\DeclareMathOperator{\tr}{Tr}
\DeclareMathOperator{\st}{subject\:to}

\def\Omegaz{{\boldsymbol{\Omega}}}
\def\Lambdaz{{\boldsymbol{\Lambda}}}

\def\Sigmaz{\boldsymbol{\Sigma}}

\newcommand{\UU}{{\mathbf U}}
\newcommand{\VV}{{\mathbf V}}
\newcommand{\WW}{{\mathbf W}}
\newcommand{\CC}{{\mathbf C}}
\newcommand{\XX}{{\mathbf X}}
\newcommand{\YY}{{\mathbf Y}}
\newcommand{\II}{{\mathbf I}}

\newcommand{\QQ}{{\mathbf Q}}

\newcommand{\PP}{{\mathbf P}}
\newcommand{\GG}{{\mathbf G}}
\newcommand{\AAA}{{\mathbf A}}

\def\uz{{\boldsymbol{u}}}

\def\xz{{\boldsymbol{x}}}
\def\yz{{\boldsymbol{y}}}

\def\0z{{\boldsymbol{0}}}

\newcommand{\cblue}{\textcolor{black}}

\begin{document}

\title{Regularized Multivariate Analysis Framework for Interpretable High-Dimensional Variable Selection}

%\author{Michael~Shell,~\IEEEmembership{Member,~IEEE,}
%        John~Doe,~\IEEEmembership{Fellow,~OSA,}
%        and~Jane~Doe,~\IEEEmembership{Life~Fellow,~IEEE}% <-this % stops a space
%\thanks{M. Shell was with the Department
%of Electrical and Computer Engineering, Georgia Institute of Technology, Atlanta,

\author{
\IEEEauthorblockN{Sergio~Mu\~noz-Romero,\thanks{Corresponding Author: Sergio Mu\~noz-Romero (e-mail: sergio.munoz@urjc.es)}}
\IEEEauthorblockA{Department of Signal Processing and Communications, Universidad Rey Juan Carlos, Madrid, SPAIN}\\
\and
\IEEEauthorblockN{Vanessa~G\'omez-Verdejo, and Jer\'onimo~Arenas-Garc\'ia,}
\IEEEauthorblockA{Department of Signal Processing and Communications,
Universidad Carlos III de Madrid, Madrid, SPAIN}
}

\maketitle

\vspace{-1cm}

\begin{abstract}
Multivariate Analysis (MVA) comprises a family of well-known methods for feature extraction which exploit correlations among input variables representing the data. One important property that is enjoyed by most such methods is uncorrelation among the extracted features. Recently, regularized versions of MVA methods have appeared in the literature, mainly with the goal to gain interpretability of the solution. In these cases, the solutions can no longer be obtained in a closed manner, and more complex optimization methods that rely on the iteration of two steps are frequently used. This paper recurs to an alternative approach to solve efficiently this iterative problem. The main novelty of this approach lies in preserving several properties of the original methods, most notably the uncorrelation of the extracted features. Under this framework, we propose a novel method that takes advantage of the $\ell_{2,1}$ norm to perform variable selection during the feature extraction process. Experimental results over different problems corroborate the advantages of the proposed formulation in comparison to state of the art formulations.
\end{abstract}

%% Note that keywords are not normally used for peerreview papers.
%\begin{IEEEkeywords}
%  Multivariate Analysis, Dimensionality Reduction, Low-Rank Solution, $\ell_{2,1}$ Norm Regularization, Feature Selection.
%\end{IEEEkeywords}

\IEEEpeerreviewmaketitle

\section{Introduction}
\begin{figure*}[t]
  \centering
     \includegraphics[width=5.5in]{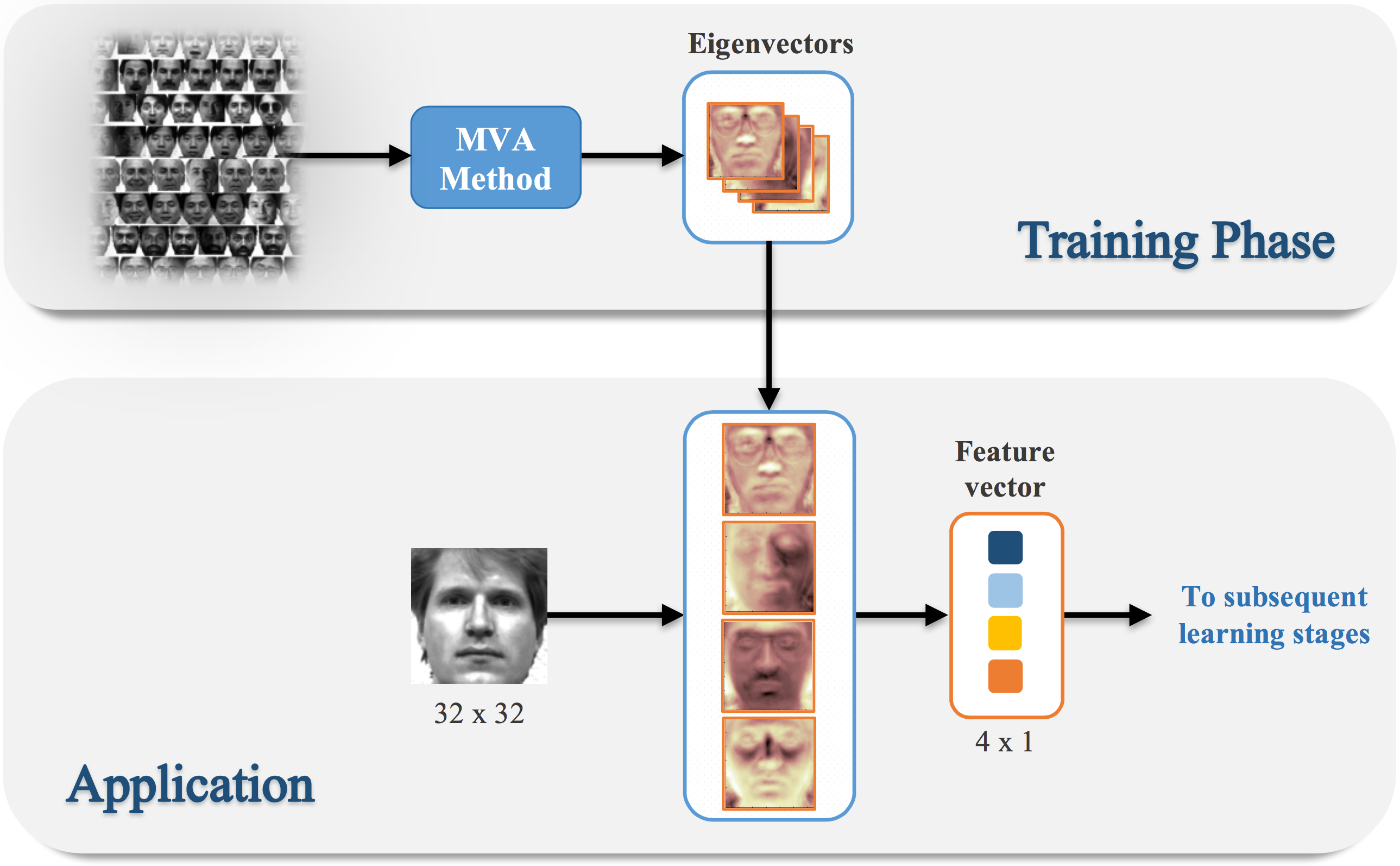}
 \caption{Feature extraction with Multivariate Analysis. During the training phase, MVA methods learn the most relevant directions for a particular dataset (in face recognition these vectors are known as {\it eigenfaces} when the PCA method is applied). Feature extraction is then carried out multiplying any vector in the input representation space with these eigenvectors. Subsequent learning tools can then be applied on this new subspace of reduced dimensionality.}
 \label{Fig:schemeMVA}
\end{figure*}

Multivariate Analysis (MVA) comprises a collection of tools that play a fundamental role in statistical data analysis. These techniques have become increasingly popular since the proposal of Principal Component Analysis (PCA) in 1901 \cite{pearson1901pca}.  PCA was proposed as a simple and efficient way to reduce data dimension by projecting the data over the largest variance directions. As illustrated in Fig. \ref{Fig:schemeMVA}, PCA learns from a given dataset a set of projection vectors, so that data can be represented in a low-dimensional space that preserves the directions of the input space where the data shows the largest variance.  A typical example to illustrate PCA is face recognition, where the projection vectors are known as eigenfaces \cite{Turk91}. Nevertheless, PCA has been used in many other applications, and can indeed be considered as one of the most widely-used tools for feature extraction.

\begin{figure*}[t]
  \begin{center}
  \begin{tabular}{cc}
%     \multicolumn{2}{c}{\includegraphics[width=\textwidth]{figures/classPCA_OPLS.pdf}}\\
%     (a) PCA & (b) OPLS
    \includegraphics[width=\textwidth]{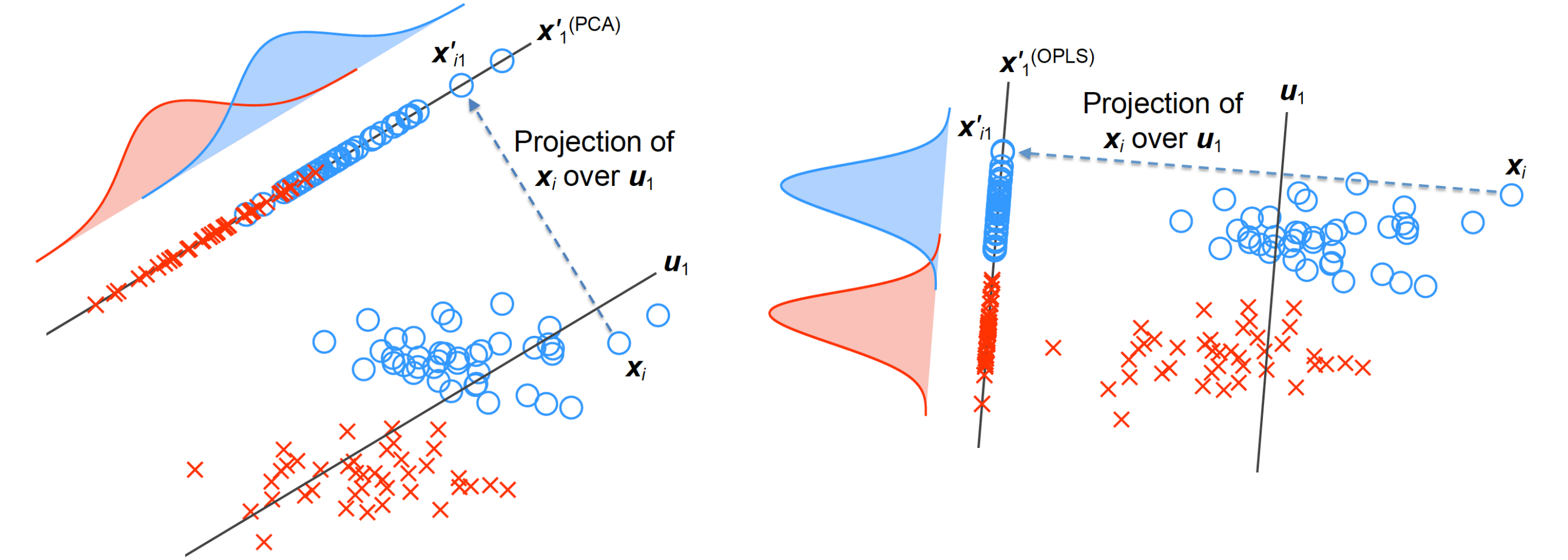}\\
    (a) PCA $\qquad\qquad\qquad\qquad\qquad\qquad\qquad\qquad\qquad\qquad\qquad$ (b) OPLS
  \end{tabular}
  \end{center}
 \caption{Projected data over the first eigenvector of PCA and OPLS in a binary classification task.}
 \label{Fig:Intro_MVA}
\end{figure*}

Other MVA algorithms have emerged that are especially suited to supervised learning tasks (e.g., in regression and classification).  In these problems, the goal is not just to represent the input data as efficiently as possible, but it actually becomes of major importance to keep the directions of the input space that are more highly correlated with the label information.  This is the case of algorithms such as Canonical Correlation Analysis (CCA) \cite{hotelling1936cca}, Partial Least Squares (PLS) approaches \cite{wold1966nipals2,wold1966nipals1}, and Orthonormalized PLS (OPLS) \cite{worsley1998mvlm}.  Consider for instance a toy classification problem in Fig. \ref{Fig:Intro_MVA}.  In this problem, the direction of the maximum variance extracted by PCA (left subplot) results in overlapping distributions of the two classes along this direction, while a supervised method like OPLS (right subplot) successfully identifies the most discriminative information.  Although this toy example is based on a classification task, the same advantages of supervised MVA over standard PCA are encountered in regression tasks---see \cite{Arenas13} for a detailed theoretical and experimental review of these methods.
%\begin{figure*}[t]
%  \centering
%  \subfloat[][PCA]{
%     \includegraphics[height=2in]{figures/classPCA3.pdf}
%     \label{Fig:projMVA}
%  }
%  ~
%  \subfloat[][OPLS]{
%     \includegraphics[height=1.6in]{figures/classOPLS3.pdf}
%     \label{Fig:performMVA}
%  }
% \caption{Projected data over the first eigenvector of PCA and OPLS in a binary classification task.}
% \label{Fig:Intro_MVA}
%\end{figure*}

The simplicity of these methods, as well as the availability of highly-optimized libraries for solving the linear algebra problems they involve, justifies the extensive use of MVA in many application fields, such as biomedical engineering \cite{Gerven12,Hansen07}, remote sensing \cite{Arenas08,Arenasbook}, or chemometrics \cite{Barker03}, among many others (see also \cite{Arenas13} for a more detailed review of application-oriented research in the field).

An important property of PCA, OPLS, and CCA is that they lead to uncorrelated variables, so that the feature extraction process provides additional advantages:
\begin{itemize}
\item The relevance of each extracted feature is directly given by the magnitude of its associated eigenvalue, which simplifies the selection of a reduced subset of features, if necessary.
\item Subsequent learning tasks are simplified, more notably, when  the covariance matrix inversion is required. This is the case of least-square based problems, such as Ridge Regression or lasso (least absolute shrinkage and selection operator) \cite{Tibshirani_94}. 
\end{itemize}

Standard versions of MVA methods implement just a feature extraction process, in the sense that all original variables are used to build the new features. However, over the last few years there have been many significant contributions to this field that have focused on gaining interpretability of the extracted features by incorporating sparsity-inducing norms, such as the $\ell_1$ and $\ell_{2,1}$ norms \cite{Nie10}, as a penalty term in the minimization problem. When these regularization terms are included, the projection vectors are favored to include zeros in some of their components, making it easier to understand the process to build the new features and thus gaining in interpretability. In fact, the $\ell_{2,1}$ rewards solutions that perform a real variable selection process, in the sense that some of the original variables are excluded from all projection vectors at once. In other words, only a subset of the original variables are used to build the new features.

Some of the most significant contributions in this direction are sparse PCA \cite{Zou06}, sparse OPLS \cite{Gerven12}, group-lasso penalized OPLS (also known as Sparse Reduced Rank Regression, SRRR) \cite{Chen12}, and $\ell_{2,1}$-regularized CCA (or L21SDA) \cite{Shi14}. 
All these approaches are based on an iterative process which combines the optimization of two coupled least-squares problems, one of them subject to a minimization constraint. Since the inspiring work \cite{Zou06}, this constrained least-squares minimization has been typically treated as an orthogonal Procrustes problem \cite{Schonemann1966}, an approach that can still be considered mainstream (see, e.g., the very recent works \cite{Lai16,Hu16}).

A first objective of this paper is to highlight and make the computational intelligence community aware of some limitations derived from the use of orthogonal Procrustes in the context of regularized MVA methods. As explained in \cite{Sergio16}, these methods
\begin{enumerate}
\item do not converge to their associated non-regularized MVA solutions when the penalty term is removed,
\item are highly dependent on initialization, and may even fail to progress towards a solution,
\item do not in general obtain uncorrelated features. 
\end{enumerate}

As solution to these problems, \cite{Sergio16} proposes an alternative optimization procedure avoiding the use of the Procrustes solution. In this paper, we will briefly review the framework presented in \cite{Sergio16} to derive regularized MVA versions, and illustrate the approach and its associated advantages by introducing a novel MVA method using the $\ell_{2,1}$ norm \cite{Nie10} as regularization term. Apart from the advantages we have already discussed implying variable selection over the original variables, this norm holds the property of rotational invariance, a fact that we will exploit to significantly reduce the computational cost of the training phase. Although some authors have already adapted the robust variable selection method \cite{Nie10} to the MVA scenario (see, e.g., the group-lasso penalized OPLS method \cite{Chen12} or the $\ell_{2,1}$-regularized CCA \cite{Shi14}), these adaptations are based on orthogonal Procrustes and the rotational invariance property of the $\ell_{2,1}$ norm is not exploited, taking unnecessary extra computational burden.

In short, the main contributions of this paper can be summarized as:
\begin{itemize}
\item Review a framework for regularized MVA methods, and explain an alternative to the most commonly used Procrustes solution to overcome the limitations of this approach.
\item Obtain novel MVA algorithms based on $\ell_{2,1}$ regularization.
\item Illustrate the effectiveness of these algorithms to carry out feature extraction and, at the same time, obtain some understanding of the original input variables.
\end{itemize}

The rest of the paper is organized as follows. Section \ref{sec:RegMVAframework} reviews the common framework for regularized multivariate analysis, and explains an advantageous alternative to the use of orthogonal Procrustes in this context. Then, Section \ref{sec:MVAmethodsL21penalty} 
particularizes the MVA framework by including an $\ell_{2,1}$ norm penalty, explaining in detail how to derive a computationally efficient solution and pointing the differences between our proposal and other existing solutions. Section \ref{sect_expe} is devoted to experiments and Section \ref{sec:Conclusions} draws the main conclusions of our work.

%: Firstly, Section 2 introduces this generalized MVA framework. Then, Section 3 presents the iterative process required to solve its regularized extension and describes both Procrustes solution, as well as our proposal based on a standard eigenvalue problem. Section 4 extends the proposed MVA framework by including an $\ell_{2,1}$ norm penalty, paying special attention to its efficient solution, taking advantage of the rotational invariance property, as well as pointing out the differences of this proposal with state of the art solutions. Section 5 illustrates and compares the suitability of the new proposed solution with that of Procrustes using some real problems which support all theoretical findings. Finally, Section 6 concludes the paper.

\section{Regularized MVA framework: enforcing feature uncorrelation}
\label{sec:RegMVAframework}

Let us assume a supervised learning scenario, where the goal is to carry out feature extraction in the input space, learning the projection vectors from a training dataset of $N$ input-output pairs $\{\xz_i,\yz_i\}_{i = 1}^N$, where $\xz_i \in \Re^n$ and $\yz_i \in \Re^m$ are the input and output vectors, respectively. Therefore, $n$ and $m$ denote the dimensions of the input and output spaces. For notational convenience, we define the input and output data matrices: $\XX = \left[\xz_1,\dots,\xz_N \right]$ and $\YY = \left[\yz_1,\dots,\yz_N \right]$, with columnwise arranged patterns. \cblue{It will be assumed throughout the paper that these matrices are centered \cite{Shawe04}}, so that sample estimations of the input and output data covariance matrices, as well as of their cross-covariance matrix, can be calculated as $\CC_{\XX\XX} = \XX\XX^\top$, $\CC_{\YY\YY} = \YY\YY^\top$ and $\CC_{\XX\YY} = \XX\YY^\top$, where we have neglected the scaling factor $\frac{1}{N}$, and superscript $^\top$ denotes vector or matrix transposition. The goal of linear MVA methods is to find $n_f$ relevant features by combining the original variables, i.e., $\XX' = \UU^\top \XX$, where the $k$th column of $\UU = [\uz_1,\dots,\uz_{n_f}]$ is a vector containing the coefficients associated to the $k$th extracted feature. % and $n_f$ is the total number of extracted features.% Similarly, some methods consider also the extraction of features for the output data. 
\cblue{Note that we are referring to the components of ${\bf x}$ as \textbf{variables}, whereas the components of ${\bf x' = {u^\top {\bf x}}}$ are being referred to as \textbf{features}.  Consequently, feature extraction implies obtaining ${\bf x'}$ from ${\bf x}$, whereas variable selection is the process of selecting a subset of the original variables in ${\bf x}$.  Besides, the feature extraction process can also imply variable selection when the projection matrix ${\UU}$ has some of their rows equal to zero.}

In this paper, we deal with MVA methods which force the extracted features to be uncorrelated; this applies, at least, to PCA, CCA, and OPLS. MVA methods that do not enforce feature uncorrelation, more notably PLS, are therefore left outside the scope of this paper.  A common framework for these regularized MVA methods can be set including an uncorrelation constraint, $\UU^\top \CC_{\XX\XX} \UU = \II$, over the formulation of \cite{reinsel98}: %According to it, these MVA framework pursues the minimization of the following objective function\footnote{This problem is different from standard least squares regression since matrix $\UU$ imposes a representation bottleneck \cite{Roweis99}.}:
\begin{align}
\label{GOPLS_cost}
{\rm  minimize_{\,\WW,\UU}~} & \quad \|\Omegaz ^{\frac{1}{2}} \left(\YY  - \WW \UU^\top \XX\right) \|_F^2 + \gamma R\left(\UU\right) \\
{\rm subject~to} & \quad \UU^\top \CC_{\XX\XX} \UU = \II \nonumber
\end{align}
where $\WW$ is an $m\times n_f$ matrix of regression coefficients, %that can alternatively be seen as a projection matrix for the output data, 
parameter $\gamma$ trades off the importance of the regularization term $R\left(\UU\right)$, $\|\AAA\|_F=\tr\{\AAA\AAA^\top\}$ denotes the Frobenius norm of matrix $\AAA$, and $\tr\{\cdot\}$ is the trace operator. Finally, different selections of matrix $\Omegaz$ give rise to the considered MVA methods, in particular $\Omegaz=\CC_{\YY\YY}^{-1}$ for CCA,  $\Omegaz=\II$ for OPLS, and $\Omegaz=\II$ with  $\YY=\XX$ for PCA \cite{reinsel98,Sergio15}.% Note that the solution to \eqref{GOPLS_cost} is not unique since, e.g., $\WW$ can compensate any scaling of matrix $\UU$. \cblue{Como ahora he incluido la restricción en el problema de optimizacion, creo que esta ultima frase ya no es cierta}

The objective function in \eqref{GOPLS_cost} is composed of two terms. The first term tries to minimize the reconstruction error when matrix $\YY$ is estimated from the projected data as $\WW\XX' = \WW\UU^\top \XX$. Note that this is different from standard least-squares since the introduction of matrix $\UU$ imposes a representation bottleneck \cite{Roweis99}, i.e., matrix $\YY$ needs to be approximated from a matrix $\XX'$ with less features than the original matrix $\XX$. The regularization term 
$R\left(\UU\right)$ is usually a particular matrix norm that gives a desired property to the solution. Three common regularization terms are:
\begin{itemize}
\item $R\left(\UU\right)=\|\UU\|_F^2=\sum_{ij}U_{ij}^2$, where $U_{ij}$ is the element in the $i$th-row and $j$th-column of $\UU$. This term is known as Tikhonov, ridge, or $\ell_2$ regularization, and it is used to improve the conditioning of the solution. %, thus making possible a direct numerical solution.
\item $R\left(\UU\right)=\|\UU\|_1=\max_j \sum_{i=1}^N |U_{ij}|$, where $|U_{ij}|$ is the absolute value of $U_{ij}$. This term is known as lasso regularization \cite{Tibshirani_94} and it is frequently used to induce sparsity on the solution matrix (i.e., to nullify some elements of $\UU$).
\item $R\left(\UU\right)=\|\UU\|_{2,1}=\sum_{i=1}^n\|\uz^i\|_2$, being $\uz^i$ the $i$th row of $\UU$. This is known as $\ell_{2,1}$ regularization and penalizes all $n_f$ coefficients corresponding to a single variable as a whole, making them drop to zero jointly, thus favoring variable selection. %\cred{Note that lasso regularization can be used for feature selection when m = 2 only (i.e., when solution is a vector instead of a matrix). In the next section, we will use this $\ell_{2,1}$ regularization term to select the most relevant variables and thus to obtain more knowledge about the problem.}
\end{itemize} 
 
\begin{figure}[t]
  \centering
  \includegraphics[width=3.45in]{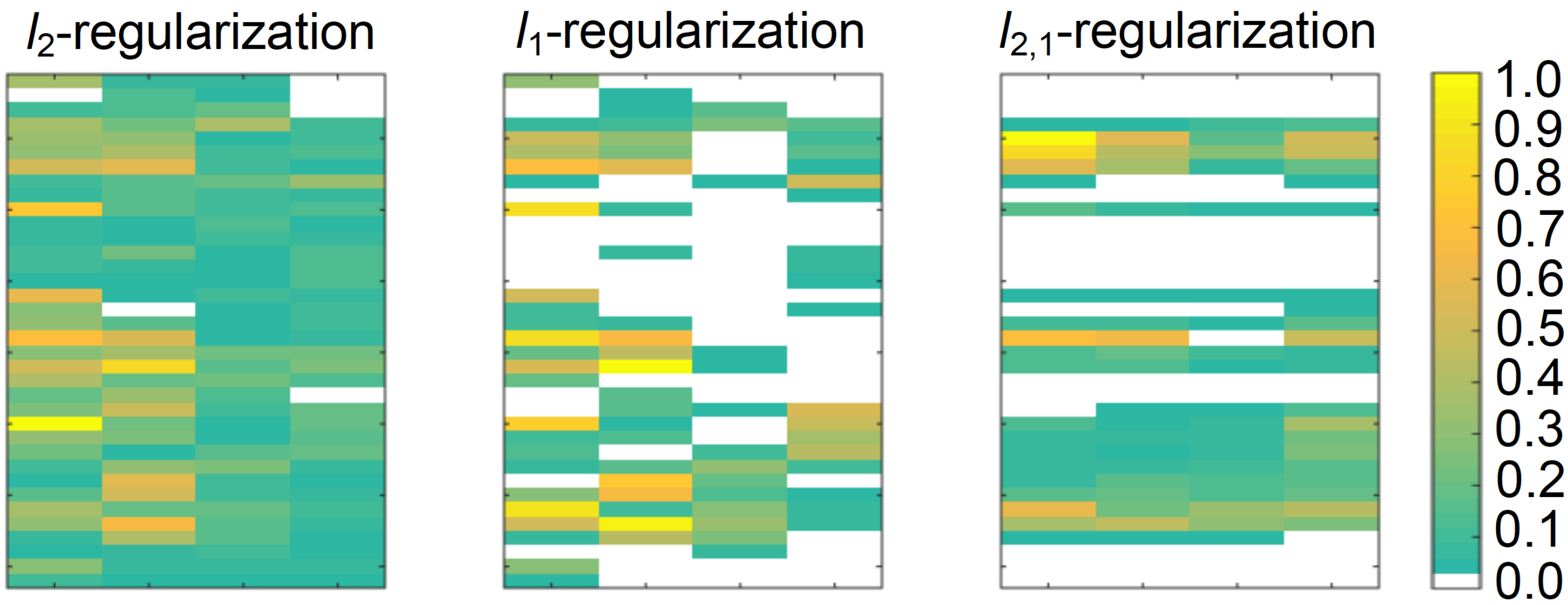}
%  \begin{tabular}{cc}
%     \includegraphics[width=3.4in]{figures/l2_l1_l21_white.pdf}  &
%     \hspace{-1cm}\includegraphics[width=.1in]{figures/colorbar_white.pdf}
%  \end{tabular}
 \caption{Examples of the provided projection matrices $\UU$ for a case with $n=40$ and $n_f = 4$ and considering the regularizations $\ell_2$, $\ell_1$, and $\ell_{2,1}$. Coefficients that take a zero value have been identified and represented in white.}
 \label{Fig:l2_l1_l21}
\end{figure}

Fig. \ref{Fig:l2_l1_l21} depicts the solution matrix $\UU$ for the above regularization terms over a toy problem. As it can be seen, %$\ell_2$ regularization doesn't provide sparsity over the coefficients of $\UU$, whereas
$\ell_1$ and $\ell_{2,1}$ penalties result in many elements of $\UU$ dropping to zero. Furthermore, $\ell_{2,1}$ norm provides the sparsity in a structured way, i.e., the coefficients of $\UU$ are annulled by rows. Since each row of $\UU$ is associated to a different input variable, this structured sparsity implies that many input variables are completely ignored during the feature extraction, so that variable selection is also achieved in this case. % a feature selection to the feature extraction process. %In the next section, we will use this regularization term to select the most relevant variables and thus to obtain more knowledge about the problem.

%When the derivatives of the regularization term, $R(\UU)$, can be calculated (for instance, when the  $\ell_2$ penalty is used), the solution of \eqref{GOPLS_cost} can be easily obtained. For this purpose, one can obtain a closed-form solution for $\UU$ as a function of $\WW$ and, then, introduce this solution back into \eqref{GOPLS_cost} to rewrite the problem in terms of $\WW$ only. 

When using non-derivable penalties, such as $\ell_1$ or $\ell_{2,1}$ norms, the solution of the minimization problem in \eqref{GOPLS_cost} cannot be obtained in closed-form. However, \cblue{as shown in \cite{Sergio15}} an equivalent formulation can be obtained by replacing the uncorrelation constraint in \eqref{GOPLS_cost} by $\WW^\top\Omegaz \WW=\II$, leading to
\begin{align}
\label{GOPLS_cost2}
{\rm minimize_{\,\WW,\UU}} & \quad \|\Omegaz ^{\frac{1}{2}} \left(\YY  - \WW \UU^\top \XX\right) \|_F^2 + \gamma R\left(\UU\right) \\
{\rm subject~to} & \quad \WW^\top\Omegaz \WW=\II. \nonumber
\end{align}

As demonstrated in \cite{Sergio15}, \eqref{GOPLS_cost} and \eqref{GOPLS_cost2} provide the same solution, but using \eqref{GOPLS_cost2} is normally more efficient for the common case $m \ll n$. Furthermore, placing the constraint on matrix $\WW$ allows to solve the problem with the two-step iterative method that we describe in Algorithm \ref{tab:UWsteps} (please, refer to \cite{Sergio15} for further details). For simplicity, the solution of the $\WW$-step has been written in terms of a new matrix $\VV=\Omegaz ^{\frac{1}{2}}\WW$. As it can be seen, the minimization problems involved by the $\UU$- and $\WW$-steps are coupled, so it becomes necessary to iterate both steps until some convergence criterion is met.

%\begin{table}
%\begin{center}
%\caption{Summary of the two steps involved in the minimization problem \eqref{GOPLS_cost2}.\label{tab:UWsteps}}
%\begin{tabular}{ll}
%\toprule
%$\UU$-Step & $\WW$-Step \\
%\midrule
%For fixed $\WW$ (satisfying $\WW^\top\Omegaz\WW=\II$): & For fixed $\UU$:\\
%$\UU^* = \argmin_{\UU}\|\YY' - \UU^\top \XX\|_F^2 + \gamma R\left(\UU\right)$ \qquad\qquad\qquad & $\VV^* = \argmin_{\VV} \|\bar\YY - \VV \XX'\|_F^2$ \\ & $\st \VV^\top\VV=\II$\\
%with $\YY'=\WW^\top\Omegaz\YY$ & with $\VV=\Omegaz ^{\frac{1}{2}}\WW$,  $\bar\YY=\Omegaz^{-\frac{1}{2}}\YY$, and $\XX' = \UU^\top \XX$\\
%\bottomrule
%\end{tabular}
%\end{center}
%\end{table}

\begin{algorithm*}[ht]
%\captionsetup{labelformat=algorithm}
%\renewcommand{\arraystretch}{1.3}
%\vspace{0.5cm}
%\caption{\cblue{Summary of the two steps involved in the minimization problem \eqref{GOPLS_cost2}.}\label{tab:UWsteps}}
\centering
\cblue{
\begin{tabular}{ll}
\toprule
1.- Input: $\XX$, $\YY$, $\Omegaz$.\\
2.- Optimization algorithm:\\
$\quad$ 2.1.- Initialize $\WW^{(0)}=\II$.\\
$\quad$ 2.2.- For $k = 1, 2, \dots$ \\
$\quad$ $\quad$ 2.2.1.- ($\UU$-Step) For fixed $\WW$ (satisfying $\WW^\top\Omegaz\WW=\II$):\\
$\qquad\qquad\qquad\quad \UU^* = \argmin_{\UU}\|\YY' - \UU^\top \XX\|_F^2 + \gamma R\left(\UU\right)$\\
$\qquad\quad\quad\quad~$ with $\YY'=\WW^\top\Omegaz\YY$.\\
$\quad$ $\quad$ 2.2.2.- ($\WW$-Step) For fixed $\UU$:\\
$\qquad\qquad\qquad\quad \VV^* = \argmin_{\VV} \|\bar\YY - \VV \XX'\|_F^2$ \\
$\qquad\qquad\qquad\quad \st \VV^\top\VV=\II$\\
$\qquad\quad\quad\quad~$ with $\VV=\Omegaz ^{\frac{1}{2}}\WW$,  $\bar\YY=\Omegaz^{-\frac{1}{2}}\YY$, and $\XX' = \UU^\top \XX$.\\
$\quad$ $\quad$ 2.2.3.- Back to 2.2.1. until convergence criterion is met. \\
3.- Output: $\UU$, $\VV$.\\
\bottomrule
\end{tabular}
}
\vspace{0.3cm}
\caption{\cblue{Summary of the two steps involved in the minimization problem \eqref{GOPLS_cost2}.}\label{tab:UWsteps}}
\end{algorithm*}

The $\UU$-step is a regularized least-squares problem that can take advantage of a great variety of existing efficient solvers \cite{Nie10,Grant08,Kim2008}. With respect to the $\WW$-step, it is important to point out that uncorrelation has to be enforced. When this is not the case, the above steps can provide infinite coupled pairs of solutions which are rotated versions of the desired ones, and the uncorrelation property of the extracted features is lost.

%\cred{At this point, it is important to point out that when this iterative procedure is used, it is not sufficient to impose $\VV^\top \VV = \II$ during the $\WW$-step to force the uncorrelation of the extracted features, in fact, this uncorrelation has to be additionally imposed during the Step-$\WW$ to actually obtain uncorrelated features. When this is not the case, the above steps can provide infinite coupled pairs of solutions $\UU'=\UU\RR$ and $\WW'=\WW\RR$, which are rotated versions of the desired ones $\UU$ and $\WW$ by multiplying by whatever rotation matrix $\RR$ ($\WW'\UU'^\top=\WW\RR\RR^\top\UU^\top=\WW\UU^\top$). This can be critical for feature extraction purposes, since when a bottle-neck is applied over $\UU$ this can result on an important lost of information.}

In fact, in the literature $\WW$-step is typically solved by using the orthogonal Procrustes approach. As it has been proved in \cite{Sergio16}, this solution neglects the uncorrelation among the extracted features. In spite of this limitation, since its initial proposal in \cite{Zou06} for the sparse PCA algorithm, it has been later extended to supervised approaches such as sparse OPLS \cite{Gerven12}, group-lasso penalized OPLS \cite{Chen12}, and $\ell_{2,1}$-regularized CCA (or L21SDA) \cite{Shi14}. In this respect, this paper aims (1) to highlight the limitations of Procrustes when used as part of the above iterative method, and (2) to encourage the adoption of an alternate method for the $\WW$-step that pursues feature uncorrelation.

\subsection{Solving the $\WW$-step with Orthogonal Procrustes}

The minimization problem in the $\WW$-step is known as orthogonal Procrustes, and its optimal solution is given by $\VV_\text{P} = \QQ\PP^\top$, where $\QQ$ and $\PP$ are the matrices of the singular value decomposition $\CC_{\bar\YY\XX'}=\QQ\Sigmaz\PP^\top$ \cite{Schonemann1966}, with $\CC_{\bar\YY\XX'}=\bar{\YY}\XX'^\top$, and where $\bar{\YY}$ was defined in Algorithm \ref{tab:UWsteps}.

However, when the Procrustes solution is adopted, the uncorrelation of the extracted features is not explicitly imposed during this step. In the simplest case when the regularization is not used ($\gamma=0$), this causes that the  extracted features differ from those of the corresponding standard MVA formulation (see \cite{Sergio16} for further details and experimental results). For the general case in which $\gamma>0$, the Procrustes solution results in higher correlation among the features than when using the alternative method described in the next subsection, as will also be illustrated in Section \ref{sect_expe}.

Apart from the correlation among the extracted features, the use of Procrustes also makes the algorithm highly dependent on initialization. For instance, it can be shown that when the regularization is removed and $\VV$ is initialized as an orthogonal matrix the algorithm fails to progress at all.

\subsection{Solving the $\WW$-step as an Eigenvalue Problem}
\label{subsec:nuestro}

In \cite{Sergio16}, it is proven that the uncorrelation of the extracted features can be obtained if the $\WW$-step is solved by means of the following eigenvalue problem:
\begin{equation}
\label{eq:reg_V}
\Omegaz^{\frac{1}{2}}\CC_{\XX\YY}^\top\UU\UU^\top \CC_{\XX\YY}\Omegaz^{\frac{1}{2}}\VV= \VV\Lambdaz,
\end{equation}
being $\Lambdaz$ the diagonal matrix containing the $n_f$ largest eigenvalues arranged in decreasing order.

The desired uncorrelation is obtained due to this eigenvalue problem forces the diagonalization of the matrix $\UU^\top\CC_{\XX\YY}\Omegaz^\frac{1}{2}\VV$, which is a necessary condition to meet the uncorrelated extracted features.

Table \ref{Tab:summary_regMVA} includes a summary of the $\UU$- and $\WW$-steps for the particular cases of regularized CCA, OPLS and PCA, when formulating the $\WW$-step as an eigenvalue problem. Remember that $\WW$ can be straightforwardly computed from $\VV$ using the relation $\WW=\Omegaz^{-\frac{1}{2}}\VV$.

\begin{table*}[!t]
\caption{Proposed solution for the two coupled steps of most popular regularized MVA methods.}
\label{Tab:summary_regMVA}
\centering
%\resizebox{\textwidth}{!}{
%\begin{footnotesize}
\begin{tabular}{@{}lll@{}}
\toprule
& $\UU$-step (reg. LS) & $\WW$-step (eigenvalue problem)\\
\midrule
reg. CCA & $\displaystyle\argmin_\UU \|\YY' - \UU^\top \XX\|_F^2 + \gamma R\left(\UU\right)$ & $\CC_{\YY\YY}^{-\frac{1}{2}}\CC_{\XX\YY}^\top\UU\UU^\top \CC_{\XX\YY}\CC_{\YY\YY}^{-\frac{1}{2}}\VV = \VV\Lambdaz$\\
reg. OPLS & $\displaystyle\argmin_\UU \|\YY' - \UU^\top \XX\|_F^2 + \gamma R\left(\UU\right)$ & $\CC_{\XX\YY}^\top\UU\UU^\top \CC_{\XX\YY}\VV = \VV\Lambdaz$\\
reg. PCA & $\displaystyle\argmin_\UU \|\XX' - \UU^\top \XX\|_F^2 + \gamma R\left(\UU\right)$ & $\CC_{\XX\XX}^\top\UU\UU^\top \CC_{\XX\XX}\VV = \VV\Lambdaz$\\
%$\WW$: & $\WW=\CC_{\YY\YY}^{-\frac{1}{2}}\VV$ & $\WW=\Gammaz^{-\frac{1}{2}}\VV$\\
\bottomrule
\end{tabular}%\end{footnotesize}%}
\end{table*}

%\subsection{Relationship between both solutions}

%In \cite{Sergio16} we also proof that, in the absence of regularization, the solution to the eigenvalue problem \eqref{eq:reg_V} is given by $\VV_\text{EIG} = \QQ$, where the columns of $\QQ$ are the left singular vectors of matrix $\CC_{\bar\YY\XX'} = \QQ \Sigmaz \PP^\top$. \cred{As expected, this implies that the solution of our method is just a rotation of the solution obtained with Procrustes, $\VV_\text{P} = \QQ \PP^\top$. However, this rotation plays a crucial role at uncorrelating the extracted features and causes an important lost of information when a bottle-neck is applied over matrix $\UU$ to obtain a reduce subset of extracted features.}

%%%%%%%%%%%%%%%%%%%%%%%%%%%%%%%%%%%%%%%%%%%%%%%%%%%%%
\section{MVA methods with $\ell_{2,1}$ penalty}
\label{sec:MVAmethodsL21penalty}

In this section, we particularize the presented MVA framework for the $\ell_{2,1}$ regularization norm. In this way, we can take advantage of the variable selection property enjoyed by this norm and obtain an algorithm that can simultaneously perform dimensionality reduction and variable selection.

For this purpose, let us replace $R\left(\UU\right)=\|\UU\|_{2,1}$ in \eqref{GOPLS_cost2}. Rewriting also the minimization problem in terms of $\VV=\Omegaz^{\frac{1}{2}}\WW$, we arrive at
%Using also , in such a way that we start minimizing the following constrained problem:
%\begin{equation*}
%\begin{array}{l}
%{\cal L}(\WW,\UU) = \|\Omegaz^{\frac{1}{2}} \left(\YY  - \WW \UU^\top \XX\right) \|_F^2 + \gamma \|\UU\|_{2,1},\vspace{3mm}\\
%\st: ~ \WW^\top\Omegaz\WW=\II,
%\end{array}
%\end{equation*}
%which, using the variable transformation $\WW=\Omegaz^{-\frac{1}{2}}\VV$, can be rewritten as
%\begin{equation}
%\label{eq:loss_function}
%\begin{array}{ll}
%{\rm minimize_{\,\VV,\UU}} & \qquad \|\YY'  - \VV \UU^\top \XX\|_F^2 + \gamma \|\UU\|_{2,1},\vspace{3mm}\\
%{\rm subject~to} & \qquad \VV^\top\VV=\II,
%\end{array}
%\end{equation}
\begin{align}
\label{eq:loss_function}
{\rm minimize_{\,\VV,\UU}} & \quad \|\YY'  - \VV \UU^\top \XX\|_F^2 + \gamma \|\UU\|_{2,1},\vspace{3mm}\\
{\rm subject~to} & \quad \VV^\top\VV=\II, \nonumber
\end{align}
where $\YY'=\Omegaz^{\frac{1}{2}} \YY$ is the new output matrix.

Considering the iterative solution detailed in Subsection \ref{subsec:nuestro}, the solution of \eqref{eq:loss_function} can be obtained by an iterative procedure consisting of two coupled steps:
%\begin{itemize}
%\item[1)] $\UU-$step: For fixed $\VV$, minimize \eqref{eq:loss_function} with respect to $\UU$ only.
%\item[2)] $\VV-$step: For fixed $\UU$, $\VV$ is obtained solving the following eigenvalue problem:
%\begin{equation}
%\label{eq:VfunctionU}
%\CC_{\XX\YY'}^\top\UU\UU^\top \CC_{\XX\YY'}\VV=\VV\Lambdaz^2,
%\end{equation}
%being $\CC_{\XX\YY'}=\CC_{\XX\YY}\Omegaz^{\frac{1}{2}}$. 
%\end{itemize}

\begin{enumerate}
\item $\UU$-step. For fixed $\VV$, find the matrix $\UU$ that minimizes the following regularized least-squares problem,
\begin{equation}
\label{eq:reg_U_L21}
\|\YY' - \VV \UU^\top \XX\|_F^2 + \gamma \|\UU\|_{2,1}.
\end{equation}
In the next subsection, we will further analyze this minimization problem to obtain an efficient solution that exploits the properties of $\ell_{2,1}$ regularization.
\item $\WW$-step. For fixed $\UU$, matrix $\VV$ is obtained solving the eigenvalue problem \eqref{eq:reg_V}.
%\begin{equation}
%\label{eq:VfunctionU}
%\CC_{\XX\YY'}^\top\UU\UU^\top \CC_{\XX\YY'}\VV=\VV\Lambdaz^2,
%\end{equation}
%being $\CC_{\XX\YY'}=\CC_{\XX\YY}\Omegaz^{\frac{1}{2}}$. 
As already discussed, existing algorithms solve this step by using orthogonal Procrustes with the undesired consequences described in previous sections. 

%approach that, as we will check in the experimental section and it has already been proved in Section \ref{section:issuesProcrustes} and \cite{Sergio15}, neglects the uncorrelation among the extracted features and can present convergence problems. 
\end{enumerate}

%S

\subsection{An Efficient Implementation for the $\UU$-step}
\label{Subsection:rotationalProperty}
To solve the $\UU$-step, we start from the iterative solution proposed in \cite{Nie10}, where $\UU$ is redefined as $\UU=\UU'\VV$ and $\UU'$ is obtained as:
 \begin{equation}\UU'=\left\lbrace \begin{array}{ll} (\CC_{\XX\XX}+\gamma\GG)^{-1}\CC_{\XX\YY'} & \text{if} ~n<N \vspace{2mm}\\ \GG^{-1}\XX(\XX^\top\GG^{-1}\XX+\gamma\II)^{-1}\YY'^\top & \text{if} ~n>N, \end{array}\right.\label{Uprimestep}\end{equation}
being $\GG$ a diagonal matrix, where its $i$th diagonal element is $G_{ii}=\frac{1}{2\|\uz^i\|_2},$ $\uz^i$ is the $i$th row of $\UU$, $n$ is the number of input variables (i.e., the number of rows of $\UU$), and $N$ is the number of training data. The straightforward application of this solution would result in a $\UU$-step involving two coupled iterative processes: one between $\UU$ and $\GG$, and other between $\UU$ and $\VV$ (note that they are coupled by means of matrix $\UU$).

However, these processes can be decoupled by taking advantage of the fact that $\VV$ is the solution of an eigenvalue problem (i.e., $\VV\VV^\top=\II$) and rewriting each diagonal term of $\GG$ as a function of $\UU'$:
\begin{equation}G_{ii}=\frac{1}{2\|\uz^i\|_2}=\frac{1}{2\|\uz'^i\VV\|_2}=\frac{1}{2 \sqrt{\uz'\VV\VV^\top\uz'^\top}}=\frac{1}{2\|\uz'^i\|_2}.
\label{eq:Gstep}
\end{equation}
In this way, the solution of the $\UU$-step is independent of matrix $\VV$. This result, known in the literature as the rotational invariance property for rows of the $\ell_{2,1}$ norm \cite{Nie10}, allows us to follow this simplified procedure: 
\begin{itemize}
\item Find the optimum $\UU'$ by iterating expressions \eqref{eq:Gstep} (for $i=1,\dots,n_f$) and \eqref{Uprimestep} until a stopping criterion is met.\footnote{\cblue{Although other criteria could be considered, we stop the method when $\tr\{\GG^{(k)}-\GG^{(k-1)}\}\leq \delta$, where the superscripts denote the iteration index and $\delta$ is a small constant, or when a maximum number of iterations have been completed.}}

\item Compute $\VV$ in a single step by solving: 
$$\CC_{\XX\YY'}^\top\UU'\UU'^\top \CC_{\XX\YY'}\VV=\VV\Lambdaz^2,$$ 
which results from \eqref{GOPLS_cost2} considering that $\UU=\UU'\VV$ and $\VV^\top\VV=\II$.
\end{itemize}
In this way, we can obtain important computational savings (as we will analyze in the experimental section). Algorithm \ref{Tab:MVAL21_pseudocode} summarizes this algorithm. This approach let us formulate $\ell_{2,1}$ based methods such as $\ell_{2,1}$-OPLS and $\ell_{2,1}$-PCA, where $\Omegaz=\II$ and the new output matrix is $\YY'=\YY$ ($\ell_{2,1}$-OPLS) or $\YY'=\XX$ ($\ell_{2,1}$-PCA); or $\ell_{2,1}$-CCA  for $\Omegaz=\CC_{\YY\YY}^{-1}$ and $\YY'=\CC_{\YY\YY}^{-\frac{1}{2}}\YY$.

\begin{algorithm*}[ht]
%\captionsetup{labelformat=algorithm}
%\renewcommand{\arraystretch}{1.3}
%\vspace{0.5cm}
%\caption{Pseudocode of MVA methods with $\ell_{2,1}$ penalty.}
\centering
%\label{Tab:MVAL21_pseudocode}
\begin{tabular}{ll}
\toprule
1.- Input: $\XX$, $\YY$, $\Omegaz$, $\gamma$.\\
2.- Optimization algorithm:\\
$\quad$ 2.1.- Initialize $\GG^{(0)}=\II$ and $\YY'=\Omegaz^{\frac{1}{2}}\YY$.\\
$\quad$ 2.2.- For $k = 1, 2, \dots$ \\
$\quad$ $\quad$ 2.2.1.- $\UU'^{(k)}=\left\lbrace \begin{array}{ll} (\CC_{\XX\XX}+\gamma\GG^{(k-1)})^{-1}\CC_{\XX\YY'} & \text{if } ~n<N \\ \GG^{(k-1)^{-1}}\XX(\XX^\top\GG^{(k-1)^{-1}}\XX+\gamma\II)^{-1}\YY'^\top & \text{if } ~n>N. \end{array}\right.$\\
$\quad$ $\quad$ 2.2.2.- $G_{ii}^{(k)}=\dfrac{1}{2\|\uz'^{i(k)}\|_2}$, for $i=1,\dots,n$.\\
$\quad$ $\quad$ 2.2.3.- If the convergence criterion is met, go to 2.3. \\
$\quad$ 2.3.- Solve eigenvalue problem $\CC_{\XX\YY'}^\top\UU'\UU'^\top \CC_{\XX\YY'}\VV=\VV\Lambdaz^2$.\\
$\quad$ 2.4.- $\UU=\UU'\Omegaz^{\frac{1}{2}}\VV$.\\
3.- Output: $\UU$, $\VV$.\\
\bottomrule
\end{tabular}
\vspace{0.3cm}
\caption{Pseudocode of MVA methods with $\ell_{2,1}$ penalty.}
\label{Tab:MVAL21_pseudocode}
\end{algorithm*}

%En el Paso 2.2.3 del correspondiente a la Tabla \ref{Tab:MVAL21_pseudocode}, se puede utilizar distintos criterios de convergencia. En el apartado de experimentos, se ha usado el mismo mecanismo de parada para todos los algoritmos: $\|\diag(\GG^{(k)}) - \diag(\GG^{(k-1)})\|_2 \leq \delta$, donde los superíndices indexan la iteración, el operador ``$\diag$'' extrae un vector con los elementos de la diagonal de la matriz correspondiente y $\delta$ es una pequeña constante. De esta manera, el algoritmo se detiene cuando las soluciones obtenidas en dos iteraciones consecutivas difieren menos de un pequeño umbral.

\subsection{Differences with State of the Art Approaches}

In principle, previously proposed L21SDA \cite{Shi14} and SRRR \cite{Chen12} algorithms attempt to solve the same problems as the algorithms $\ell_{2,1}$-CCA and $\ell_{2,1}$-OPLS presented in this paper. However, due to the procedure followed by their resolution, these state of the art algorithms suffer from the following important inconveniences: 
\begin{itemize}
\item First, they present the aforementioned drawbacks of all Procrustes based MVA methods.
\item Second, they do not exploit the rotational invariance property resulting in considerably larger computational burden in comparison with our proposal. Whereas our proposed solution completely decouples both iterative procedures and gets to reduce them to just one iterative process, where $\VV$ can be computed at the end, L21SDA method has to obtain the value of $\VV$ inside the iterative procedure. The case of SRRR algorithm is even worse, since it does not merge the two iterative processes, causing a much more expensive solution. The following section will analyze these issues over some real problems.
\end{itemize}

\section{Experiments}
\label{sect_expe}
This section analyzes the advantages of the proposed $\ell_{2,1}$-MVA framework from different points of view. To this purpose, we have split it into three subsections so that each one focuses on a different advantage of our proposal. The first subsection shows the advantages of including $\ell_{2,1}$ regularization to provide MVA methods with variable selection capabilities. Then, we will analyze the ability of the MVA approaches and, in particular, the $\ell_{2,1}$-MVA methods for dealing with data that have high multicollinearity among the input variables. This is a difficult situation for MVA methods, since linear dependency among the input variables can cause large fluctuations in the solution. Finally, we will show the importance of avoiding the Procrustes solution by comparing our proposals with state of the art L21SDA and SRRR. %For this purpose, we will both analyze the problems caused by using Procrustes based approaches in terms of uncorrelation of the extracted features, and measure the computational savings achieved by our methods due to exploiting the rotational invariance property.

%of the $\ell_{2,1}$ norm. To do this, we have selected two classification problems with high dimensionality and multicollinearity among their variables, which are summarized in Table \ref{Tab:datasets_cap7}. \cblue{In adition, we also include a toy problem to validate the advantage of extract uncorrelated features when there exist multicollinearity in a regression problem against only perform feature selection.}

%\cblue{The way of illustrating all of the following comparisons is to evaluate the performance achieved by an SVM classifier from the variables selected by the proposed methods and by the state-of-the-art algorithms.}
%: \textit{Carcinomas}, with 139 and 35 training and test samples, and 9182 and 11 input and output variables; and \textit{Yale}, with 120 and 45 training and test samples (8 images per person), and 1024 and 15 input and output variables.

\subsection{Variable Selection by means of $\ell_{2,1}$ Regularization}
\label{subsec:varSelec_l21reg}

In this section we are going to deal with a hyperspectral image segmentation and classification problem. This data set comes from a set of hyperspectral sensors mounted on satellite or airborne platform which acquire the reflected energy by the Earth with high spatial detail and in several wavelengths. In particular, we have selected the standard Airborne Visible/Infrared Imaging Spectrometer image taken over Northwest Indiana’s Indian Pine in June 1992 \cite{Pines13}. This dataset consists of 220 spectral bands, with 20 noisy bands covering the region of water absorption. Discriminating among the major crop classes in the area can be very difficult (in particular, given the moderate spatial resolution of 20 m), making the scene a challenging benchmark to validate classification accuracy of hyperspectral imaging algorithms. Besides, the large number of narrow spectral bands induce a high collinearity among variables, making MVA approaches a powerful tool for this application.

The selected hyperspectral image has $145 \times 145$ pixels and contains $17$ quite unbalanced classes (ranging from 20 to 10,776 pixels). Among the available 21,025 labeled pixels, $70\%$ were used for training the feature extractors and classifiers, and the remaining $30\%$ were taken apart for testing purposes. The discriminative power of all extracted features was tested using a linear SVM classifier.

\begin{figure*}[t]
  \centering
     \includegraphics[width=6.5in]{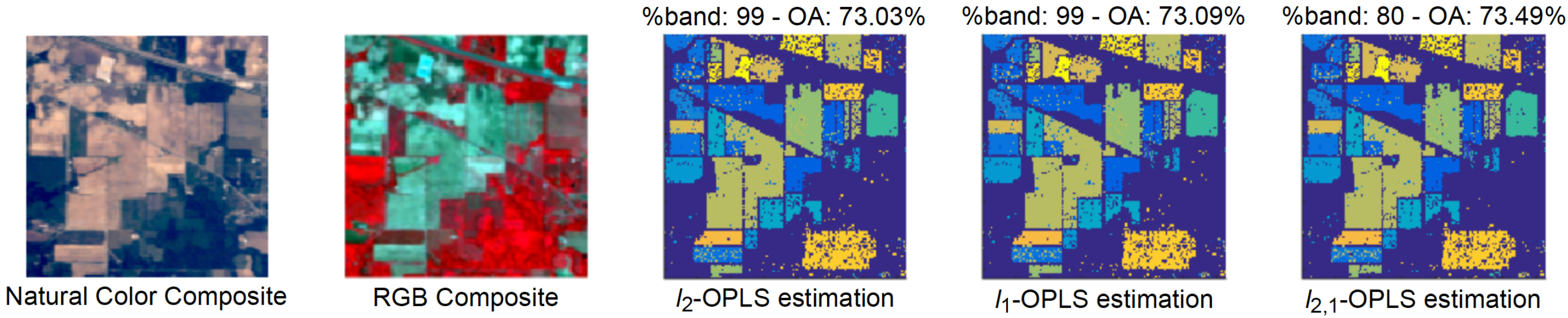}
 \caption{Natural and RGB composite (by using 50, 27 and 17 channels) hyperspectral images and its reconstructed images using $\ell_{2}$-OPLS, $\ell_{1}$-OPLS and $\ell_{2,1}$-OPLS algorithms.}
 \label{Fig:reconstrucciones}
\end{figure*}

To analyze the advantages of the $\ell_{2,1}$ penalty as variable selection tool, we are going to analyze the performance of our proposed regularized MVA framework when different regularizations are used: ridge norm or $\ell_{2}$ penalty, lasso regularization or $\ell_{1}$ norm, and the $\ell_{2,1}$ penalty. For this first study, only OPLS methods have been considered. Fig. \ref{Fig:reconstrucciones} shows the reconstructed image or the classification map for these three regularized versions of OPLS, including its overall accuracy (OA) over the test data and the percentage of selected bands ($\%$ band). In this case, regularization parameters, as well as the number of selected variables of the $\ell_{2,1}$-OPLS method, were adjusted using five-fold cross-validation in the training set.  In particular, we have explored a rectangular grid taking values from the sets $\{10^{-6}, 5\cdot 10^{-6}, 10^{-5}, 5\cdot 10^{-5}, \ldots, 50, 100, 500, 1000\}$ and $\{1, 10, 100, 1000\}$ for $\gamma$ and the SVM regularization parameter, respectively.  We have checked that these intervals are sufficiently large to ensure that the limits were not selected as a result of the CV. After this validation process, all methods show similar accuracy, with $\ell_{2}$ and $\ell_{1}$ OPLS versions achieving an accuracy of $73\%$ using almost all bands (their percentage of selected bands is around $99\%$). The performance of the $\ell_{2,1}$-OPLS method, is only slightly better with an accuracy of $73.5\%$, although it uses only  $80\%$ of the spectral bands.

 %with five fold cross validation selected parameter $C=1$.

%Figure 2 shows the test classification accuracy for a varying number of extracted features, $n_f$. For linear models, OPLS performs better than all other methods for any number of extracted features. Even though CCA provides similar results for $n_f = 10$, it involves a slightly more complex generalized eigenproblem. When the maximum number of projections is used, all methods result in the same error. Nevertheless, while PCA and PLS2 require 200 features (i.e., the dimensionality of the input space), CCA and OPLS only need 15 features to achieve virtually the same performance.
%
%The classification maps obtained for nf =10 confirm these conclusions: higher accuracies lead to smoother maps and smaller error in large spatially homogeneous vegetation covers.

\begin{figure*}[t]
  \centering
  \includegraphics[width=.75\textwidth]{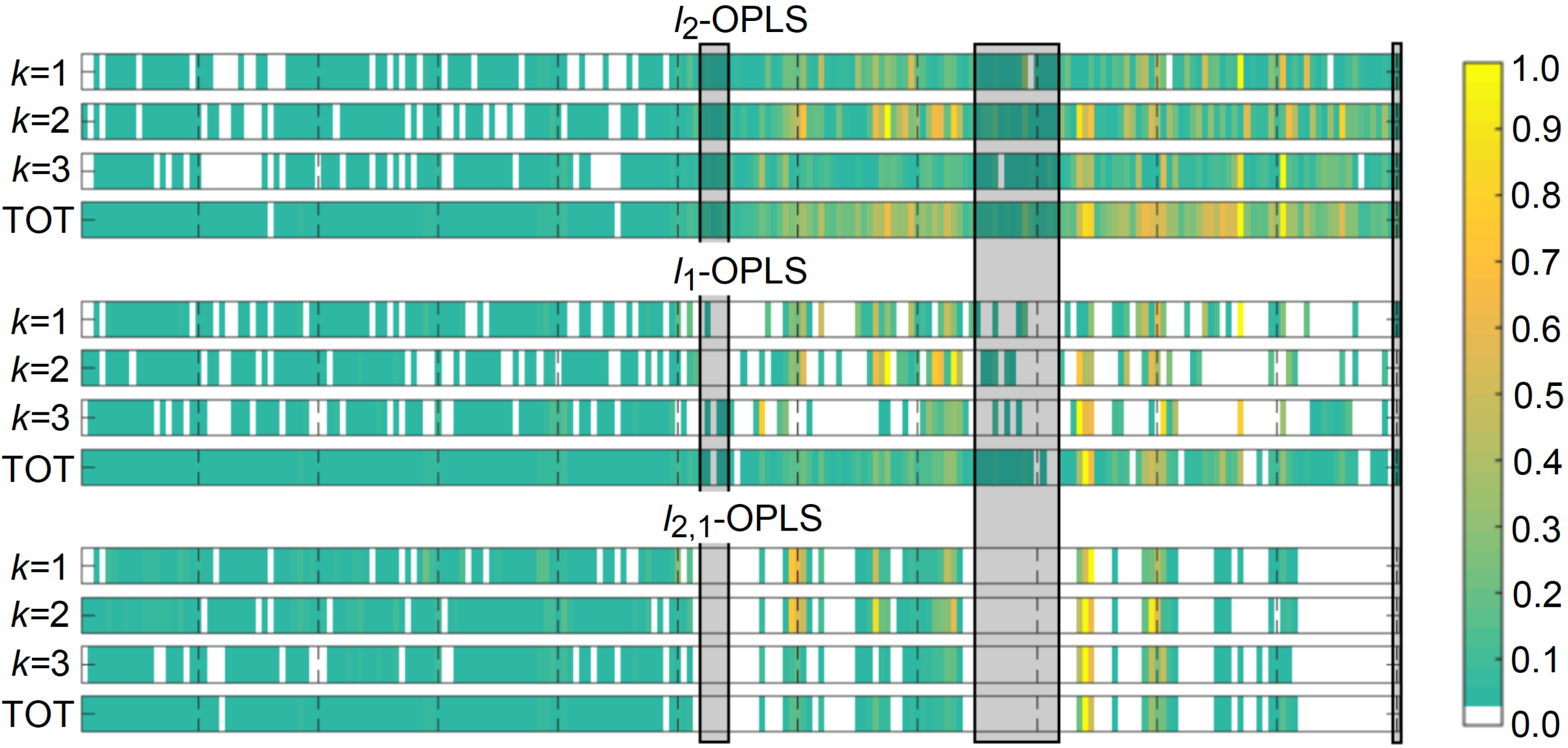}
%  \begin{tabular}{lr}
%     \includegraphics[width=4.5in]{figures/eigs_OPLS.pdf} &
%     \includegraphics[width=.16in]{figures/colorbar_white.pdf}
%  \end{tabular}
 \caption{Analysis of the band selection capabilities for $\ell_2$-OPLS, $\ell_1$-OPLS and $\ell_{2,1}$-OPLS algorithms. For each algorithm, each row plots the relevance of the components of the first three eigenvectors and the last row includes the average relevance over all eigenvectors (TOT). Water absorption bands, that should be discarded for the classification tasks, have been highlighted in grey color.}
 \label{Fig:L2_vs_L21_pines}
\end{figure*}

In order to go deeper into the structured sparsity obtained by $\ell_{2,1}$-OPLS, and its variable selection capabilities, Fig. \ref{Fig:L2_vs_L21_pines} depicts, for the three regularized OPLS versions, the importance of each variable (calculated as $\|\uz^i\|^2$) over their first three eigenvectors ($k=1,2,3$). The total importance of each variable, given as its averaged importance over all the eigenvectors, is included at the last row of the plot for each algorithm and denoted as ``TOT''. In this case, to better analyze the sparsity properties of the different regularization terms, the regularization parameter ($\gamma$) has been fixed in such a way the both $\ell_{1}$ and $\ell_{2,1}$ norms provide similar number of zeros over all the projection vectors. Furthermore, for comparison purposes, all components with a relevance value lower than $10^{-4}$ have been drop to zero. As expected, $\ell_1$-OPLS is able to nullify some of the eigenvector components and, in a few cases, this provides a variable selection (a $6\%$ of the bands have zero in all their associated components); however,  $\ell_{2,1}$-OPLS presents this sparsity in a structured way (all columns are zero), causing $35\%$ of the input bands are not used by their associated eigenvectors. Regarding $\ell_2$-OPLS, it is important to remark that it also seems to provide sparsity over its solutions, but this is mainly due to the $10^{-4}$ threshold applied over the relevant values; even so, it only removes $1\%$ of the input bands. Furthermore, $\ell_{2}$ and $\ell_{1}$ OPLS versions are selecting some positions of the noisy water bands (marked with three shadow rectangles) in the figure, whereas $\ell_{2,1}$-OPLS is able to remove all of them.

\begin{figure*}[t]
  \centering
  \includegraphics[width=.85\textwidth]{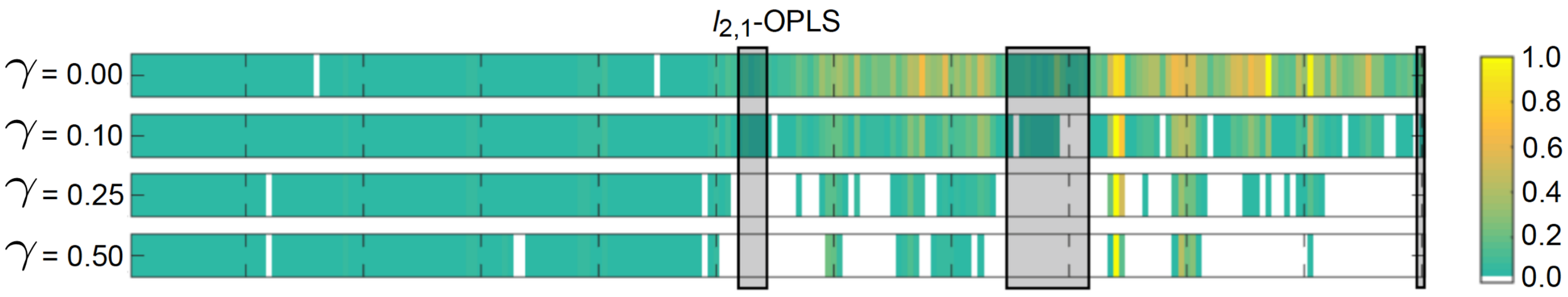}
%  \begin{tabular}{lr}
%     \includegraphics[width=4.5in]{figures/gammas_OPLS.pdf} &
%     \includegraphics[width=0.145in]{figures/colorbar_white_small.png}
%  \end{tabular}
 \caption{Analysis of the band selection capabilities for the $\ell_{2,1}$-OPLS method according to penalization term value.}
 \label{Fig:MVA_vs_L21MVA_pines}
\end{figure*}

Finally, to analyze the influence of the regularization term over the number of selected bands, Fig. \ref{Fig:MVA_vs_L21MVA_pines} displays the total importance of each variable (averaged over all eigenvectors) for $\ell_{2,1}$-OPLS  when the parameter $\gamma$ is varied from $0$ to $0.5$. When $\gamma=0$ (lack of regularization), the method is recovering standard OPLS solution, which is quite similar to $\ell_{2}$ versions in terms of sparsity. However, as larger gamma values are considered less bands are used. In particular, values of $\gamma$ close to $0.25$ are enough to remove all water absorption bands.

%\vspace{-2mm}
\subsection{Dealing with Multicollinearity of the Input Variables}
%\vspace{-2mm}
%The aim of this subsection is to illustrate the superiority of applying mostly uncorrelated feature extraction when there exist multicollinearity.

The aim of this subsection is to illustrate the advantages of combining the variable selection and feature extraction processes when there exist multicollinearity among the input variables. This is a difficult situation for MVA methods since highly correlated variables can cause large fluctuations in the solution in response to small changes in the model or data.

For this purpose, we compare the proposed methods against the state-of-the-art Robust Feature Selection (RFS)  algorithm \cite{Nie10}, which is an efficient and an outlier-robust implementation of the least squares problem with an $\ell_{2,1}$ penalization term. We start this study with a toy regression problem and, next, we carry out a similar evaluation over two classification problems with high dimensionality and multicollinearity among their variables. The main characteristics of these problems are summarized in Table \ref{Tab:datasets_cap7}.

\begin{table}[!t]
\caption{Main properties of the datasets: number of training ($N_{train}$) and test ($N_{test}$) samples, number of input ($n$) and output ($m$) variables, and number of training images per person ($p$).}
\centering
\label{Tab:datasets_cap7}
\begin{tabular}{@{}llll@{}}
\toprule
\hspace{0.3cm} & $N_{train} / N_{test}$ \hspace{0.3cm} & $n$ \hspace{0.3cm} & $m$\\ 
\midrule
%\textit{arrhythmia} \hspace{0.3cm} & 315 / 135 \hspace{0.3cm} & 276 \hspace{0.3cm} & 16\\
\textit{Carcinomas} \hspace{0.3cm}& 139 / 35 \hspace{0.3cm}& 9182 \hspace{0.3cm}& 11 \\
\textit{Yale} ($p=8$) \hspace{0.3cm}& 120 / 45 \hspace{0.3cm}& 1024 \hspace{0.3cm}& 15 \\
\bottomrule
\end{tabular} 
\end{table}

\subsection*{Regression Toy Problem with High Multicollinearity}

This toy problem consists of a simple artificial regression problem with three types of input variables: relevant, redundant and noisy. In particular, this problem considers $n = 4000$ random variables where $n_{relev} = 500$ are the relevant ones, which are generated following a Gaussian distribution with zero mean and variance randomly selected from 0 to 4; $n_{redund} = \dfrac{n}{2} = 2000$ variables can be considered redundant, since they are  obtained as a linear combination of the relevant ones; the model also includes $n_{noisy} = 1500$ noisy variables, generated as independent Gaussian variables with zero mean and unit variance. Therefore, defining the observation $\xz = \left(\xz_{relev}^\top,\xz_{redund}^\top,\xz_{noisy}^\top\right)$ and the output vector $\yz \in \mathbb{R}^{m\times 1}$, $m = 10$ being  the number of output variables, the regression model is given by:
$$\yz = \left(\begin{array}{cc} \WW_{relev} & \0z \\ \0z & \0z \end{array}\right) \xz + \epsilon,$$
where $\epsilon$ is a vector of Gaussian noise with mean 0 and variance $10^{-6}$, $\WW_{relev} \in \mathbb{R}^{m\times n_{relev}}$ is a fixed matrix with random elements selected from an uniform distribution between $-1$ and $+1$, and $\0z$ is a zero-matrix with the appropriate size. Thus, the regression coefficient matrix is built such that $\yz$ depends only on the relevant input variables.

Following the above model, we build a set of $N = 500$ training samples and apply a 70/30 ($\%$) partitioning to obtain the training and test sets, respectively. Then, we normalize both sets to zero mean and unitary standard deviation. This process is repeated over 10 random executions, obtaining independent datasets, to average the final results over these runs.

Variable selection is carried out taking the best $n_s < n$ variables after sorting them by relevance according to the corresponding values of $\|\uz'_i\|$ or $\|\uz_i\|$ (with $i=1,\dots,n$) for RFS and the $\ell_{2,1}$-MVA methods, respectively. Once the $n_s$ variables have been obtained, an optimal Least Squares (LS) regression model is adjusted by using as inputs either the $n_s$ selected original variables (in the case of RFS) or the $n_f$ features extracted from the $n_s$ variables selected by $\ell_{2,1}$-MVA algorithms. The iterative process of $\ell_{2,1}$-MVA methods is stopped when a maximum of 50 iterations are reached or when the Frobenius norm of the difference between the solutions obtained in two consecutive iterations is less than a tolerance value $\delta=10^{-6}$.

\begin{figure*}[t]
  \centering
  \subfloat[][$\gamma=0.5$]{
     \includegraphics[width=2.5in]{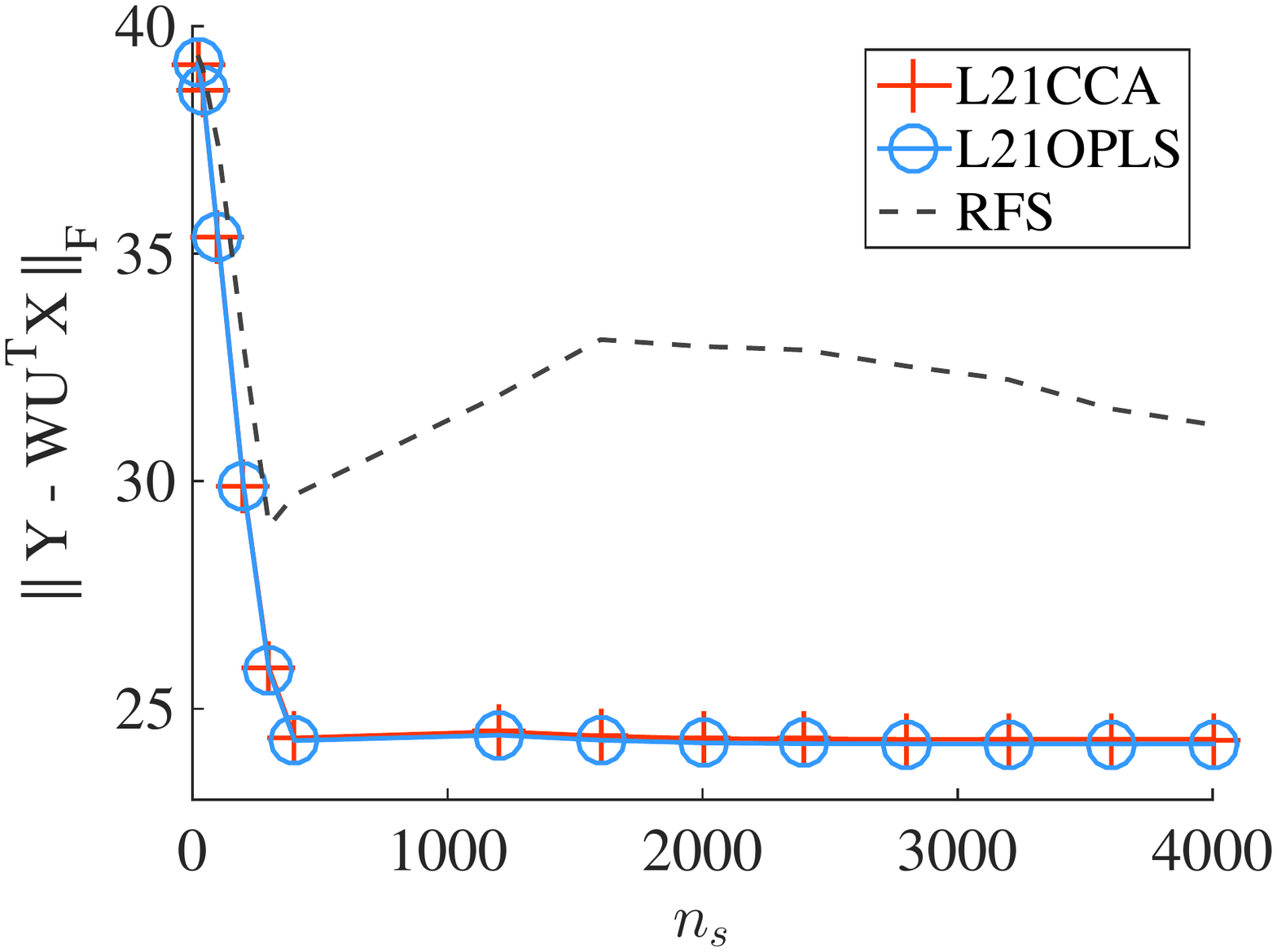}
     \label{Fig:gamma05}
  }
  ~
  \subfloat[][$\gamma=100$]{
     \includegraphics[width=2.5in]{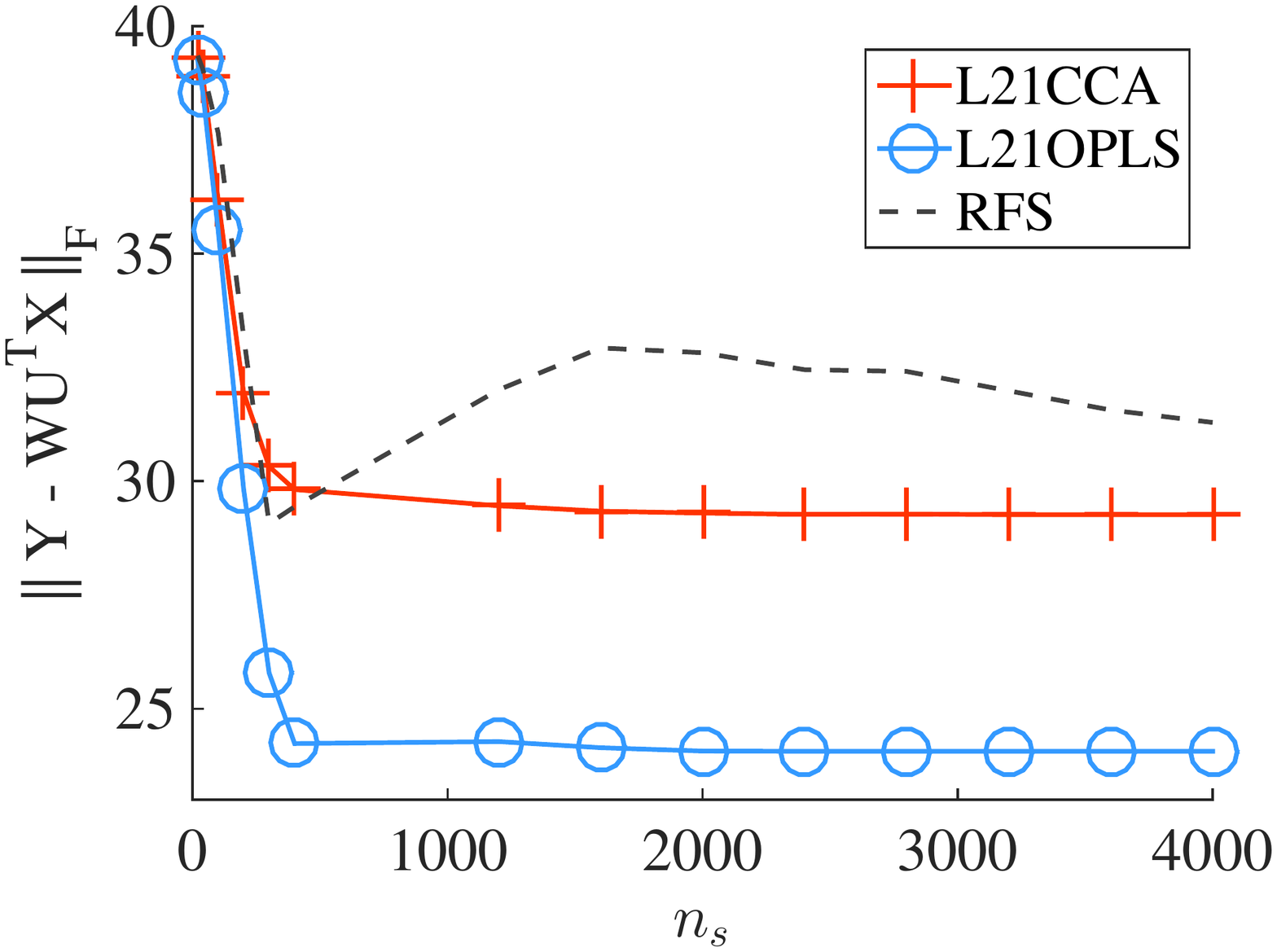}
     \label{Fig:gamma100}
  }
 \caption{Comparative curves in terms of MSE according to the number of selected variables ($n_s$) for (a) $\gamma = 0.5$ and (b) $\gamma=100$.}
 \label{Fig:ToyProblem}
\end{figure*}

In Fig. \ref{Fig:ToyProblem}, MSE obtained by the proposed $\ell_{2,1}$-CCA and $\ell_{2,1}$-OPLS algorithms using all extracted features and the reference algorithm RFS are shown according to the number of selected variables for two values of the penalty parameter: (a) $\gamma = 0.5$ selected by cross-validation, and (b) $\gamma=100$ selected to illustrate the robustness of $\ell_{2,1}$-OPLS with respect to the selection of this parameter. As can be seen, multicollinearity can cause serious problems of overfitting, in fact, this is the case of the RFS method that, although it is a robust method in the presence of outliers, suffers from serious overfitting caused by redundant variables of the problem. On the contrary, MVA methods can successfully deal with such problems.
We can see that they improve on RFS performance for $n_s<500$, and remain stable after that point with no significant degradation. It can be shown that for $n_s=500$ our methods successfully identify the relevant variables in all cases. In addition to this, $\ell_{2,1}$-MVA extracted features remain mostly orthogonal, as a consequence of the proposed optimization method.

\subsection*{Real World Classification Problems}

This subsection analyzes the advantages of the proposed $\ell_{2,1}$-MVA methods, in comparison to RFS method, over two real classification problems with high dimensionality and multicollinearities among their variables: \textit{Carcinomas} and \textit{Yale}.

To make a fair comparison between the methods under study, the free parameter of these models ($\gamma $) is selected through a 10 fold cross-validation. For this study, $\ell_{2,1}$-MVA methods use all the extracted features with the corresponding selected variables. The extracted features (both from $\ell_{2,1}$-MVA methods and RFS) are then fed to a linear SVM whose accuracy is used to evaluate the performance of each method. We explored the same ranges of values for $\gamma$ and SVM regularization parameter C as we did in \ref{subsec:varSelec_l21reg}.

\begin{figure*}[t]
  \centering
  \subfloat[][\textit{Carcinomas}]{
     \includegraphics[width=2.6in]{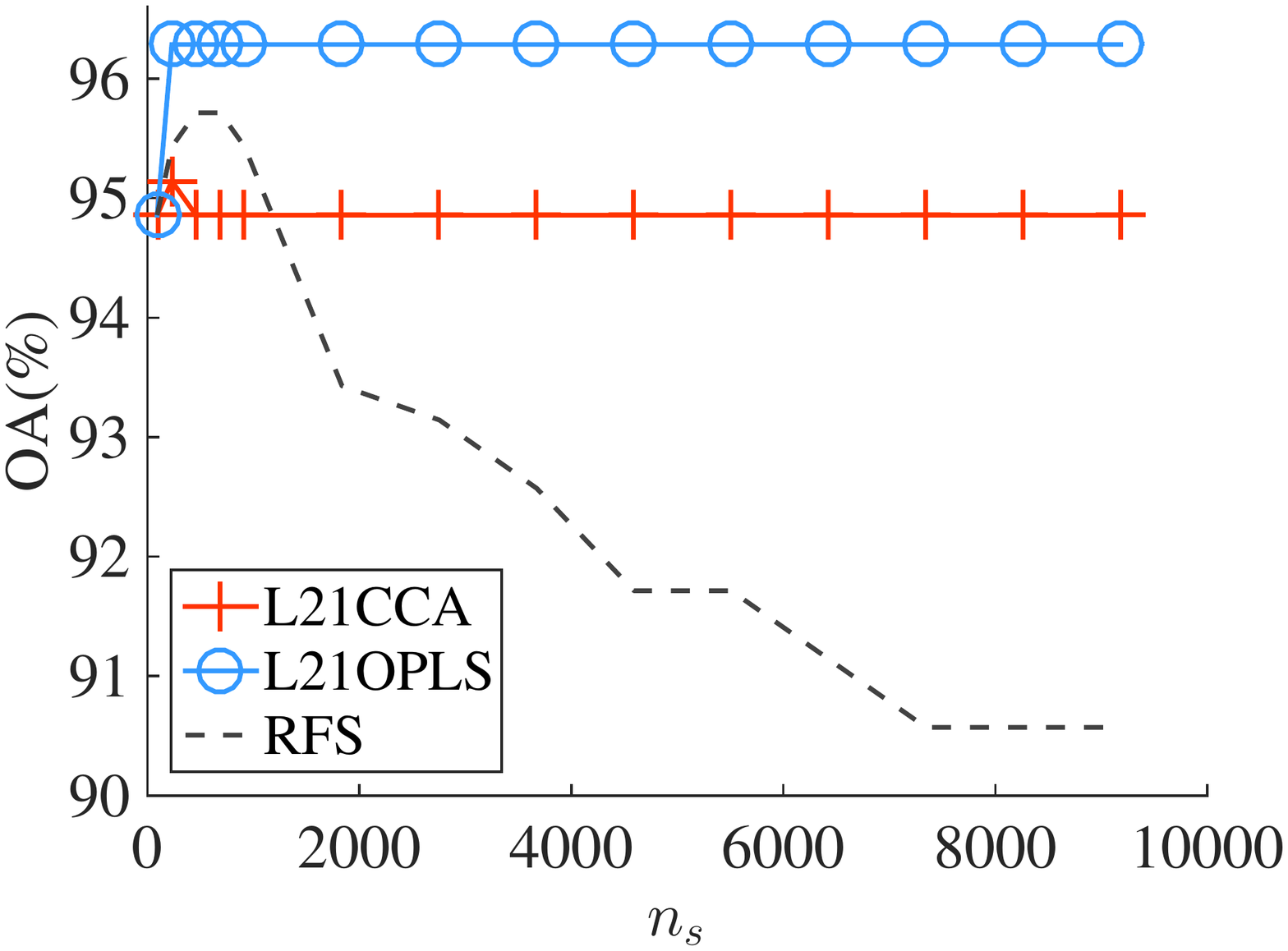}
     \label{Fig:Carcinomas}
  }
  ~
  \subfloat[][\textit{Yale}]{
     \includegraphics[width=2.6in]{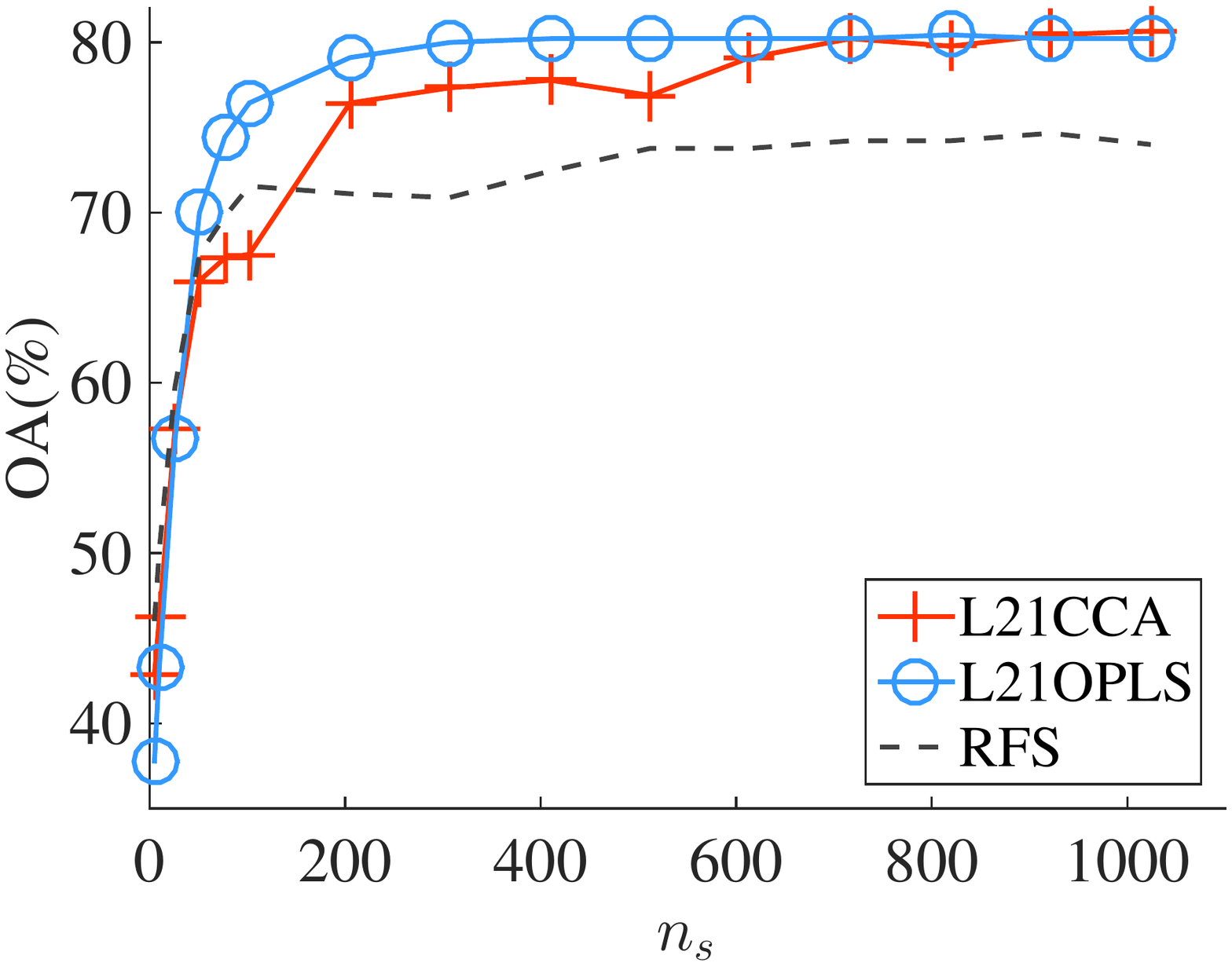}
     \label{Fig:Yale}
  }
 \caption{Comparative curves in terms of overall accuracy as a function of the number of selected variables  ($n_s$).}
 \label{Fig:OA_FS}
\end{figure*}

Fig. \ref{Fig:OA_FS} displays the overall accuracy (OA) as a function of the number of selected variables ($n_s$) for $\ell_{2,1}$-OPLS, $\ell_{2,1}$-CCA and RFS. These results corroborate the conclusions derived from the toy problem, that is, multicollinearity among variables causes RFS to suffer from overfitting problems, which is more evident in the \textit{Carcinomas} dataset; on the contrary, $\ell_{2,1}$-MVA approaches overcome this drawback. It is also interesting to see that $\ell_{2,1}$-OPLS clearly outperforms the other methods. According to the $\ell_{2,1}$-OPLS and $\ell_{2,1}$-CCA curves, one might conclude that all relevant information of the \textit{Carcinomas} dataset lies within 2$\%$ of the variables, which is where these algorithms reach their maximum performance.

%\vspace{-2mm}

\subsection{Comparison with L21SDA and SRRR}

Whereas the previous subsection compared our approaches with a pure variable selection method, in this subsection, we carry out a comparison against the state of the art approaches based on the Procrustes solution. Remember that SRRR \cite{Chen12} can be seen as an OPLS with $\ell_{2,1}$ penalty, whereas L21SDA \cite{Shi14} is a CCA version including the same penalty term. To the best of our knowledge, no PCA method with $\ell_{2,1}$ penalty has appeared in the literature to date, but its derivation following a Procrustes formulation is straightforward, and is considered here for the sake of completeness.

%In particular, we compare the proposed L21CCA and L21OPLS with L21SDA and SRRR, since they can be seen as the same algorithms respectively, except for the following two major differences: L21SDA and SRRR use the Procrustes approach to obtain matrix $\VV$ and they do not exploit the rotational invariance property of the $\ell_{2,1}$ norm. 

%\subsection*{Advantages for avoiding the usage of the Procrustes approach in the solution}
%\vspace{-2mm}

\begin{figure*}[t]
  \centering
  \subfloat[][\textit{Carcinomas}]{
     \includegraphics[width=2.5in]{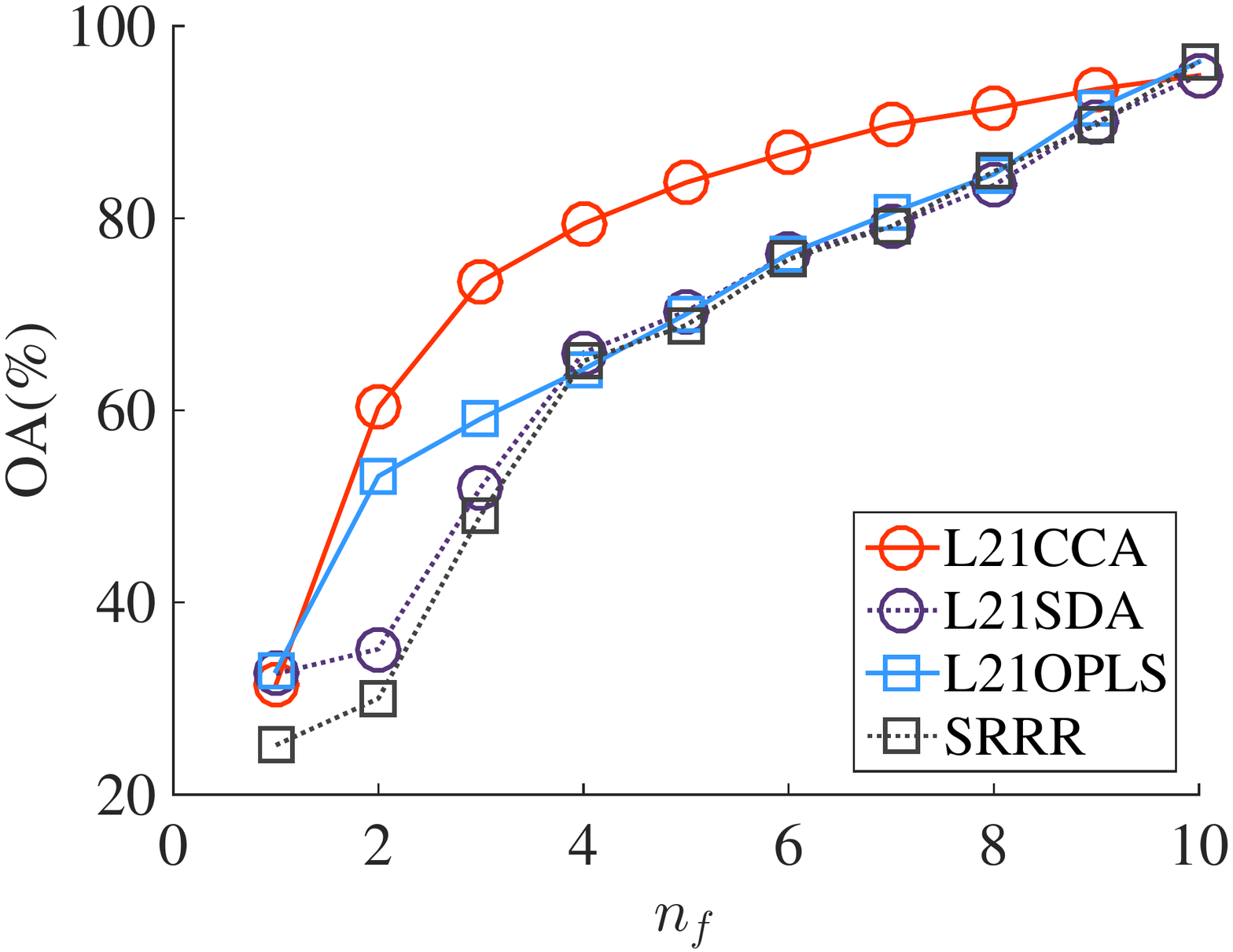}
     \label{Fig:Carcinomas_Procrustes}
  }
  ~
  \subfloat[][\textit{Yale}]{
     \includegraphics[width=2.5in]{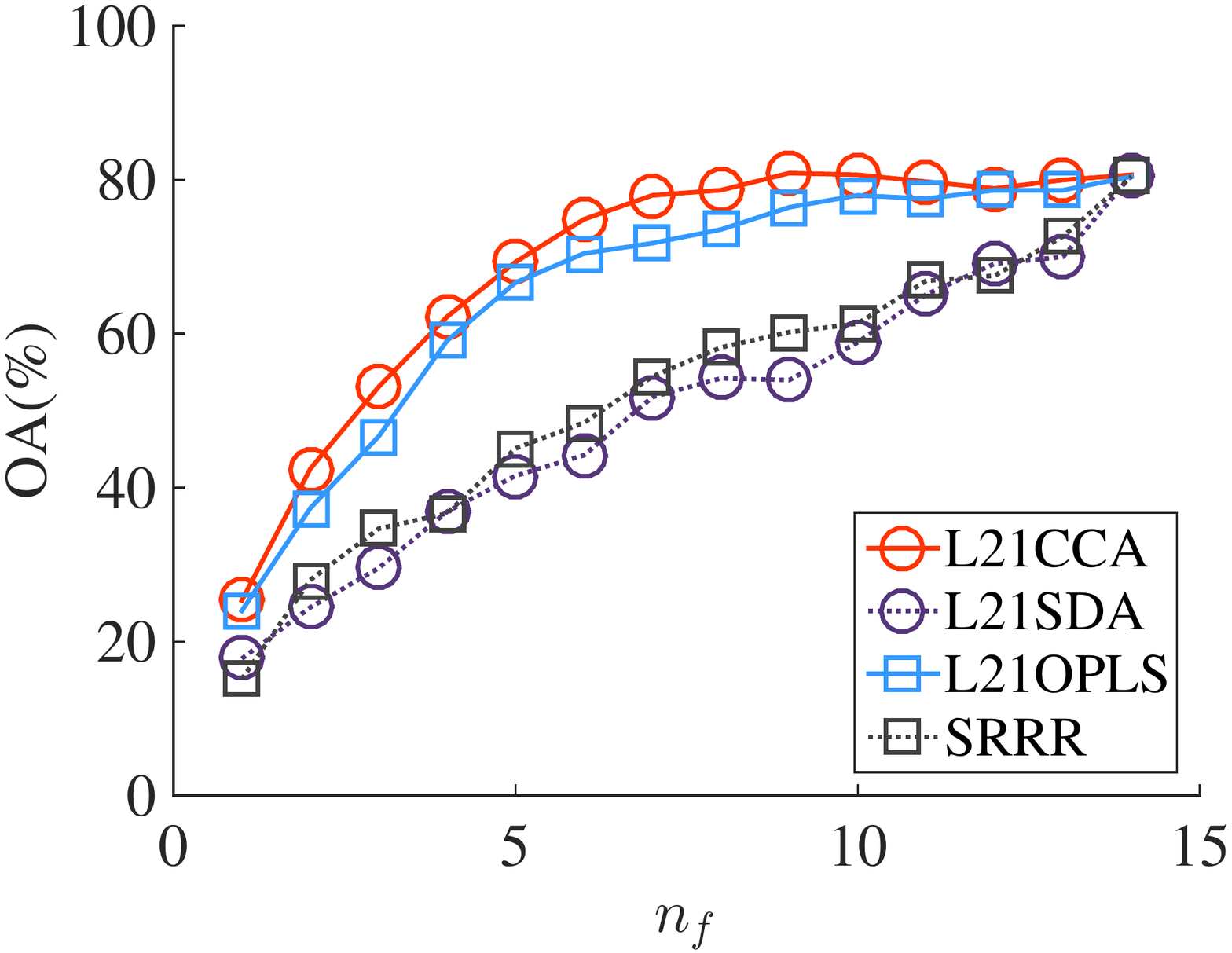}
     \label{Fig:Yale_Procrustes}
  }
 \caption{Comparative curves in terms of OA according to the number of extracted features ($n_f$) for $\ell_{2,1}$-CCA, $\ell_{2,1}$-OPLS and reference methods L21SDA and SRRR.}
 \label{Fig:Procr_FS}
\end{figure*}

Here, the experimental procedure is the same as in the previous subsection, but the curves shown below are made based on the number of features extracted instead of the number of selected variables. Fig. \ref{Fig:Procr_FS} shows the OA obtained according to the number of extracted features. When a low number of features is extracted ($n_f'<n_f$),  $\ell_{2,1}$-CCA and $\ell_{2,1}$-OPLS clearly outperform L21SDA and SRRR methods. This advantage is due to the ability of the proposed framework to extract a set of mostly uncorrelated features, making easier the training of the subsequent classifier and straightforward the selection of an optimum reduced subset of features.

\begin{figure}[t]
  \centering
       \includegraphics[width=3.4in]{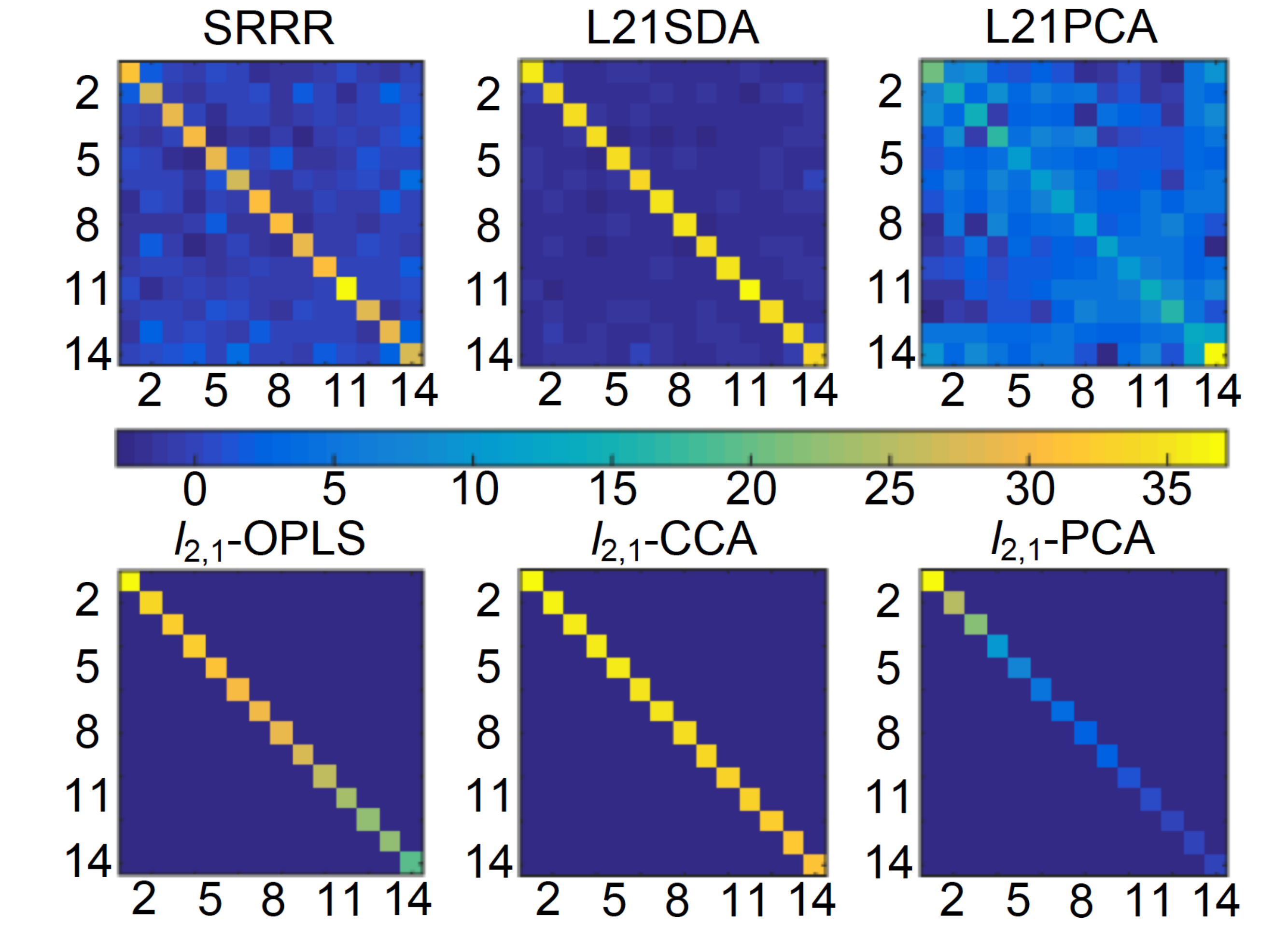}
 \caption{Comparison among the feature correlation matrices of our proposed methods and those proposed in the literature, which use Procrustes approach for the \textit{Carcinomas} dataset.}
 \label{Fig:uncorrelatedFeatures}
\end{figure}

\begin{figure*}[t]
  \centering
  \subfloat[][L21CCA]{
     \includegraphics[width=2.2in]{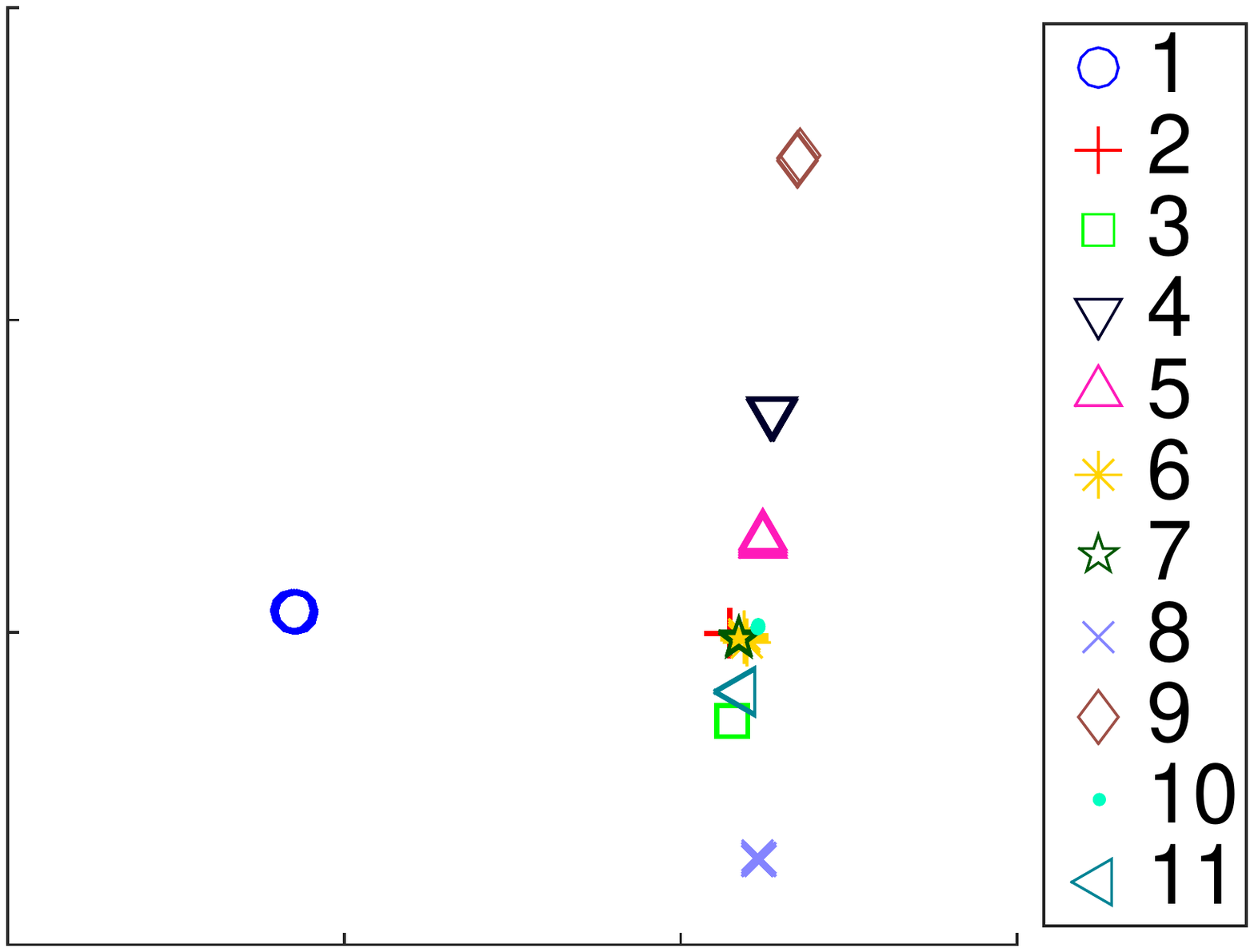}
     \label{Fig:Carcinomas_L21CCA}
  }
  ~
  \subfloat[][L21SDA]{
     \includegraphics[width=2.2in]{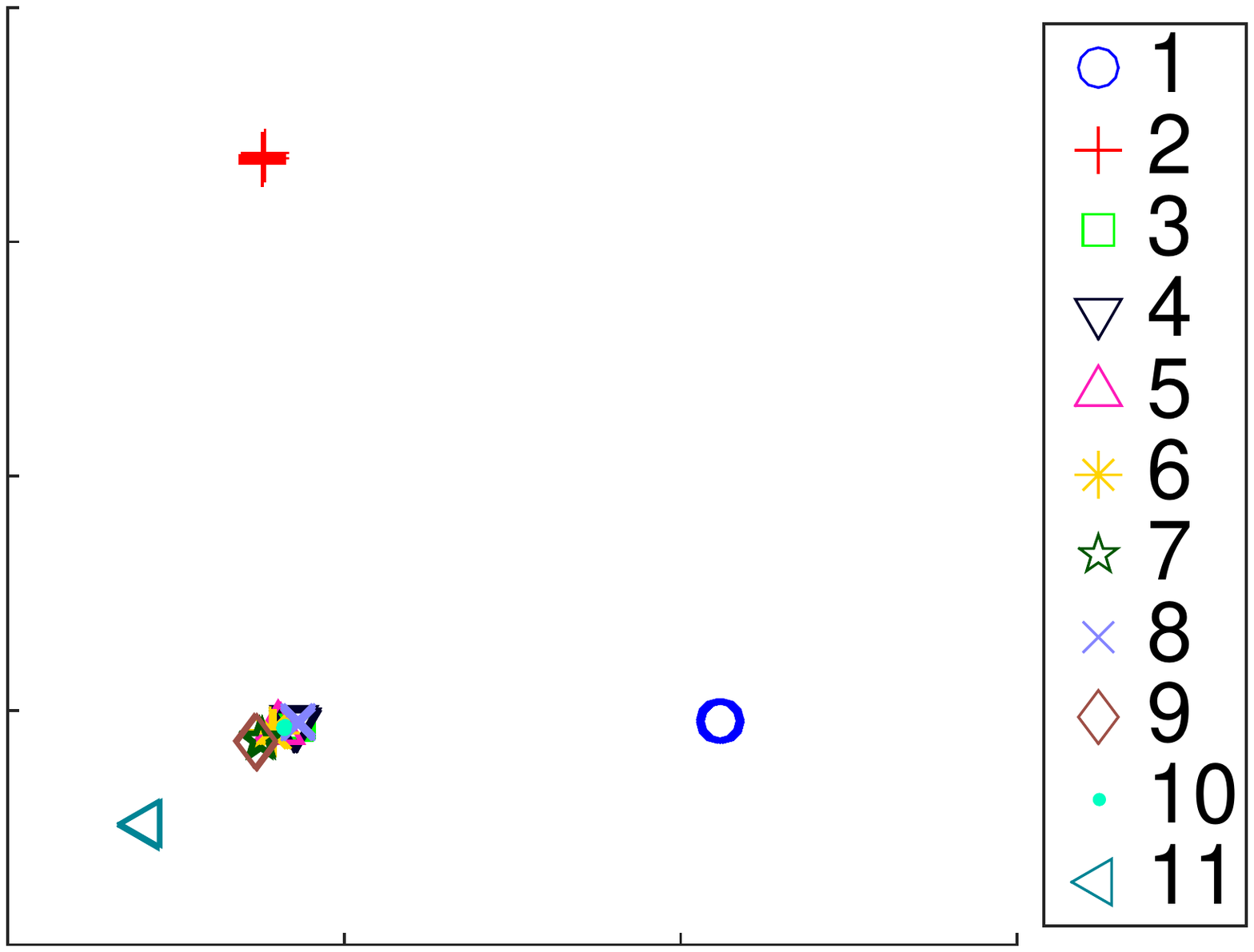}
     \label{Fig:Carcinomas_L21SDA}
  }
 \caption{Discriminative power comparison between $\ell_{2,1}$-CCA and L21SDA when only using the two first extracted features in \textit{Carcinomas} dataset.}
 \label{Fig:twoFirstProjections}
\end{figure*}

To analyze in detail this issue, Figs. \ref{Fig:uncorrelatedFeatures} and \ref{Fig:twoFirstProjections} show the correlation matrices of he projected data and the discrimination capabilities of these methods for Carcinomas dataset. As expected, the proposed $\ell_{2,1}$-MVA approaches, unlike Procrustes-based schemes, are able to obtain an almost complete uncorrelation among the projected data (note that non-diagonal terms are almost null in our proposed methods). This fact directly provides an improvement of the discrimination capability of new projected features, as Fig. \ref{Fig:twoFirstProjections} reveals. In this plot, we can check that $\ell_{2,1}$-CCA algorithm is able to project the data into a two dimensional space without overlapping among classes, making easier the subsequent classification task (in this case, the OA is close to 60$\%$); whereas, L21SDA projects most of the classes over the same region and, therefore, the classifier accuracy is reduced by half (OA is around 30$\%$).

Note that, when all the extracted features are used, the results are the same, since the final classifier (SVM) uses all the projected information, which is just a reconstruction from the original space.

\begin{figure*}[t]
  \centering
  \subfloat[][\textit{Carcinomas}]{
     \includegraphics[width=2.5in]{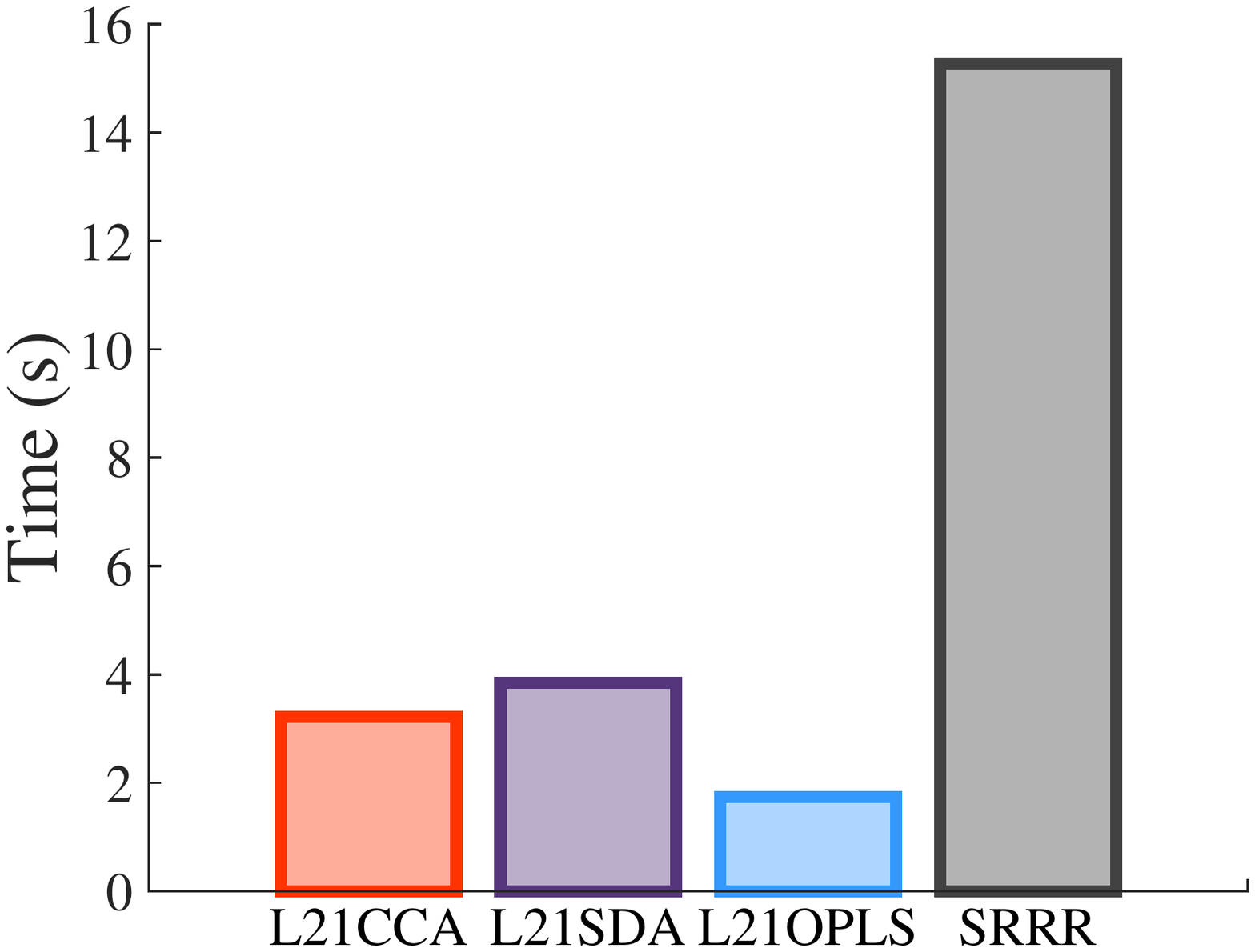}
     \label{Fig:time_Carcinomas}
  }
  ~
  \subfloat[][\textit{Yale}]{
     \includegraphics[width=2.5in]{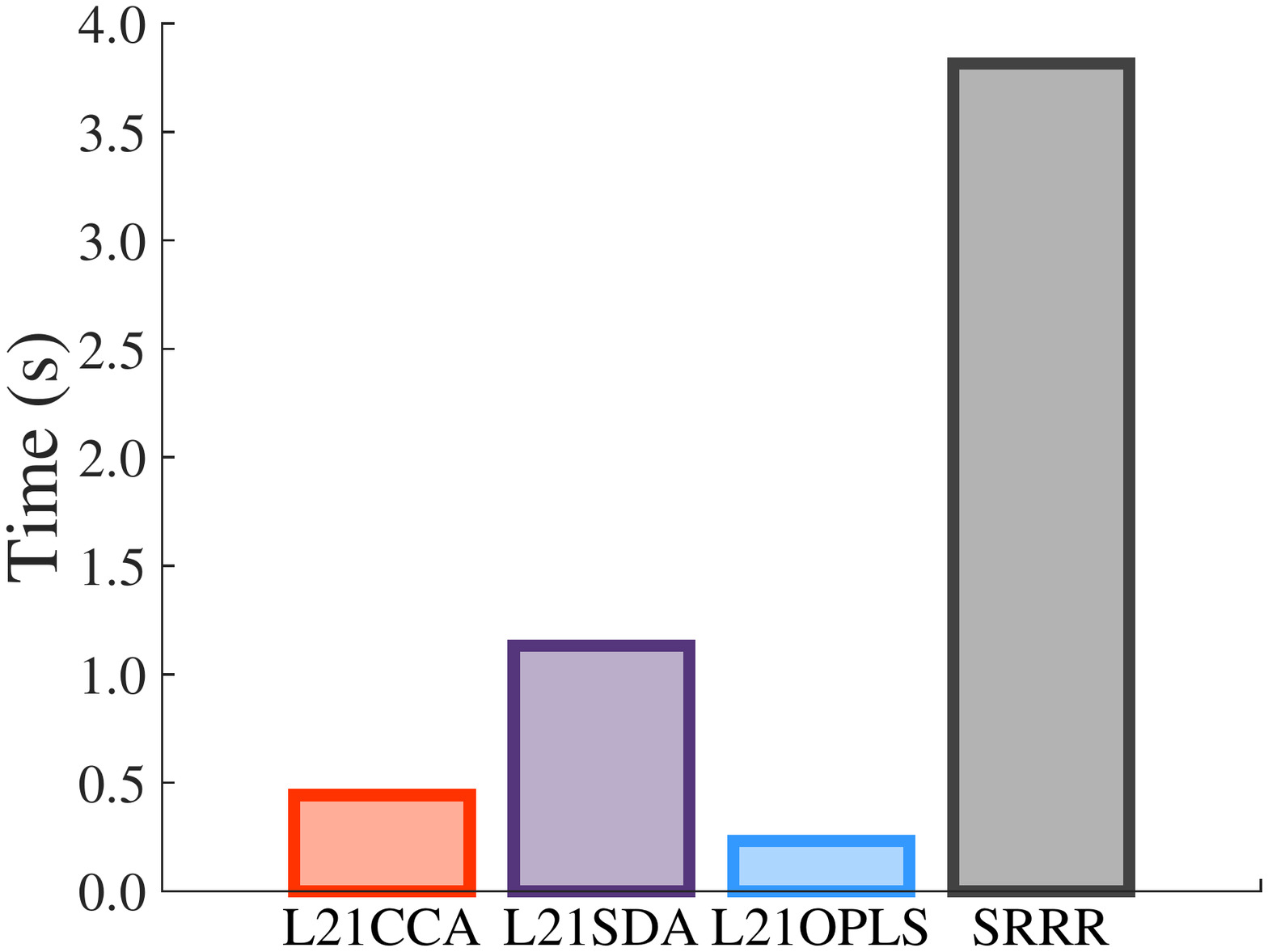}
     \label{Fig:time_Yale}
  }
 \caption{Training time (in seconds) required by $\ell_{2,1}$-CCA, $\ell_{2,1}$-OPLS, L21SDA and SRRR algorithms.}
 \label{Fig:time_cost}
\end{figure*}

%\vspace{-2mm}
%\subsection*{Computational savings through exploiting the rotational invariance of the $\ell_{2,1}$ norm}
%\vspace{-2mm}
Finally, a comparative study of the computational burden is also shown in Fig. \ref{Fig:time_cost}. As expected, proposed methods are computationally more efficient, as a direct consequence of exploiting the rotational invariance of the $\ell_{2,1}$ norm, as explained in Subsection \ref{Subsection:rotationalProperty}.%, requiring lower time to compute their solution. It specially points out the computational savings when L21OPLS is compared to SRRR: 2'' vs. 16'' and 0,25'' vs. 3,7'' in \textit{Carcinomas} and \textit{Yale}, respectively.

%It can be seen that the computational savings of L21OPLS against SRRR increase with the number of input variables (see subfigure \ref{Fig:time_cost} \subref{Fig:time_Carcinomas}). Besides, when the difference between the number of input variables and output decreases, the runtime difference between L21CCA and L21SDA increase (see Subfigure \ref{Fig:time_cost} \subref{Fig:time_Yale}).

%Therefore, we can conclude that the proposed methods, in addition to having better performance than the state-of-the-art approaches when a subset of features is selected, are computationally more efficient, since they require less time to obtain the solution.

\section{Conclusions}
\label{sec:Conclusions}

Solutions of regularized MVA approaches are based on an iterative approach consisting of two coupled steps. Whereas the first step eases the inclusion of regularization terms, the second results in a constrained minimization problem which has been typically solved as an orthogonal Procrustes problem. Despite the extended use of this scheme, it fails in obtaining a new subspace of uncorrelated features, this being a desired property of MVA solutions. In this paper we have analyzed the drawbacks of these schemes, recurring to an alternative to the Procrustes solution for the second step, that forces uncorrelation among the extracted features, and thus overcome the drawbacks of previous schemes. %The advantages of the proposed technique have been discussed theoretically, and further confirmed via simulations.

% and experimentally corroborate that Procrusters based MVA approaches, unlike proposed solution, present the following drawbacks: obtain  (in general) correlated features, do not converge to their associated standard MVA problem and its final solution is highly dependent algorithm initialization. Furthermore, experiments reveal that, since Procrusters based solutions do not force uncorrelation, the obtained regularized solutions result in higher correlation among the features than those of the proposed method.

In order to show the practical advantages of our regularized MVA solution, this paper particularizes the proposed method to derive MVA methods implementing $\ell_{2,1}$ regularization. These proposed $\ell_{2,1}$-MVA methods provide an efficient selection of the relevant variables of the problem exploiting the rotational invariance of the $\ell_{2,1}$ norm. At the same time, they can deal with the multicollinearity problems using feature extraction and providing mostly uncorrelated features. 

Finally, experimental results over high dimensional problems show that the methods included in this MVA framework are not only computationally more efficient than previous state of the art solutions, but also can improve their performance.% when there is multicollinearity among input variables. %In addition, issues have been confirmed using the orthogonal Procrustes problem in iterative schemes for parsimonious MVA solutions.

\section*{Acknowledgment}

%This work has been partly supported by MICYT project TEC2014-52289-R.
This work has been partly supported by MINECO projects TEC2013-48439-C4-1-R, TEC2014-52289-R and TEC2016-75161-C2-2-R,  and Comunidad de Madrid projects PRICAM P2013/ICE-2933 and S2013/ICE-2933.

%\bibliographystyle{IEEEtran}
%\bibliography{IEEEabrv,biblio}

\begin{thebibliography}{10}
\providecommand{\url}[1]{#1}
\csname url@samestyle\endcsname
\providecommand{\newblock}{\relax}
\providecommand{\bibinfo}[2]{#2}
\providecommand{\BIBentrySTDinterwordspacing}{\spaceskip=0pt\relax}
\providecommand{\BIBentryALTinterwordstretchfactor}{4}
\providecommand{\BIBentryALTinterwordspacing}{\spaceskip=\fontdimen2\font plus
\BIBentryALTinterwordstretchfactor\fontdimen3\font minus
  \fontdimen4\font\relax}
\providecommand{\BIBforeignlanguage}[2]{{%
\expandafter\ifx\csname l@#1\endcsname\relax
\typeout{** WARNING: IEEEtran.bst: No hyphenation pattern has been}%
\typeout{** loaded for the language `#1'. Using the pattern for}%
\typeout{** the default language instead.}%
\else
\language=\csname l@#1\endcsname
\fi
#2}}
\providecommand{\BIBdecl}{\relax}
\BIBdecl

\bibitem{pearson1901pca}
K.~Pearson, ``On lines and planes of closest fit to systems of points in
  space,'' \emph{Philosophical Magazine}, vol.~2, no.~6, pp. 559--572, 1901.

\bibitem{Turk91}
M.~Turk and A.~Pentland, ``Eigenfaces for recognition,'' \emph{Journal of
  cognitive neuroscience}, vol.~3, no.~1, pp. 71--86, 1991.

\bibitem{hotelling1936cca}
H.~Hotelling, ``Relations between two sets of variates,'' \emph{Biometrika},
  vol.~28, pp. 321--377, 1936.

\bibitem{wold1966nipals2}
H.~Wold, ``Estimation of principal components and related models by iterative
  least squares,'' in \emph{Multivariate Analysis}.\hskip 1em plus 0.5em minus
  0.4em\relax Academic Press, 1966, pp. 391--420.

\bibitem{wold1966nipals1}
------, ``Non-linear estimation by iterative least squares procedures,'' in
  \emph{Research Papers in Statistics}.\hskip 1em plus 0.5em minus 0.4em\relax
  Wiley, 1966, pp. 411--444.

\bibitem{worsley1998mvlm}
K.~Worsley, J.~Poline, K.~Friston, and A.~Evans., ``Characterizing the response
  of {PET} and {fMRI} data using multivariate linear models {(MLM)},''
  \emph{Neuroimage}, vol.~6, pp. 305--319, 1998.

\bibitem{Arenas13}
J.~Arenas-Garc{\'\i}a, K.~B. Petersen, G.~Camps-Valls, and L.~K. Hansen,
  ``Kernel multivariate analysis framework for supervised subspace learning: A
  tutorial on linear and kernel multivariate methods,'' \emph{IEEE Signal
  Processing Magazine}, vol.~30, no.~4, pp. 16--29, July 2013.

\bibitem{Gerven12}
M.~A.~J. van Gerven, Z.~C. Chao, and T.~Heskes, ``On the decoding of
  intracranial data using sparse orthonormalized partial least squares,''
  \emph{Journal of Neural Engineering}, vol.~9, no.~2, pp. 26\,017--26\,027,
  2012.

\bibitem{Hansen07}
L.~K. Hansen, ``Multivariate strategies in functional magnetic resonance
  imaging,'' \emph{Brain and Language}, vol. 102, no.~2, pp. 186--191, 2007.

\bibitem{Arenas08}
J.~Arenas-Garc{\'\i}a and G.~Camps-Valls, ``Efficient kernel orthonormalized
  {PLS} for remote sensing applications,'' \emph{{IEEE} Trans. Geosci. Remote
  Sens.}, vol.~44, pp. 2872--2881, 2008.

\bibitem{Arenasbook}
J.~Arenas-Garc{\'\i}a and K.~B. Petersen, ``Kernel multivariate analysis in
  remote sensing feature extraction,'' in \emph{Kernel Methods for Remote
  Sensing Data Analysis}, G.~Camps-Valls and L.~Bruzzone, Eds.\hskip 1em plus
  0.5em minus 0.4em\relax Wiley, 2009.

\bibitem{Barker03}
M.~Barker and W.~Rayens, ``Partial least squares for discrimination,''
  \emph{Journal of Chemometrics}, vol.~17, no.~3, pp. 166--173, 2003.

\bibitem{Tibshirani_94}
R.~Tibshirani, ``Regression shrinkage and selection via the lasso,''
  \emph{Journal of the Royal Statistical Society, Series B (Methodological)},
  vol.~58, no.~1, pp. 267--288, 1996.

\bibitem{Nie10}
F.~Nie, H.~Huang, X.~Cai, and C.~Ding, ``Efficient and robust feature selection
  via joint $\ell_{2,1}$-norms minimization,'' in \emph{Advances in Neural
  Information Processing Systems 23}.\hskip 1em plus 0.5em minus 0.4em\relax
  The MIT Press, 2010, pp. 1813--1821.

\bibitem{Zou06}
H.~Zou, T.~Hastie, and R.~Tibshirani, ``Sparse principal component analysis,''
  \emph{Journal of Computational and Graphical Statistics}, vol.~15, no.~2, pp.
  265--286, 2006.

\bibitem{Chen12}
L.~Chen and J.~Z. Huang, ``Sparse reduced-rank regression for simultaneous
  dimension reduction and variable selection,'' \emph{Journal of the American
  Statistical Association}, vol. 107, no. 500, pp. 1533--1545, 2012.

\bibitem{Shi14}
X.~Shi, Y.~Yang, Z.~Guo, and Z.~Lai, ``Face recognition by sparse discriminant
  analysis via joint ${L}_{2,1}$-norm minimization,'' \emph{Pattern
  Recognition}, vol.~47, no.~7, pp. 2447--2453, 2014.

\bibitem{Schonemann1966}
P.~H. Sch{\"o}nemann, ``A generalized solution of the orthogonal procrustes
  problem,'' \emph{Psychometrika}, vol.~31, no.~1, pp. 1--10, 1966.

\bibitem{Lai16}
Z.~Lai, W.~K. Wong, Y.~Xu, J.~Yang, and D.~Zhang, ``Approximate orthogonal
  sparse embedding for dimensionality reduction,'' \emph{IEEE Transactions on
  Neural Networks and Learning Systems}, vol.~27, no.~4, pp. 723--735, April
  2016.

\bibitem{Hu16}
Z.~Hu, G.~Pan, Y.~Wang, and Z.~Wu, ``Sparse principal component analysis via
  rotation and truncation,'' \emph{IEEE Transactions on Neural Networks and
  Learning Systems}, vol.~27, no.~4, pp. 875--890, April 2016.

\bibitem{Sergio16}
S.~Mu{\~n}oz-Romero, V.~G{\'o}mez-Verdejo, and J.~Arenas-Garc{\'\i}a, ``Why
  (and how) avoid orthogonal procrustes in regularized multivariate analysis,''
  \emph{arXiv preprint, arXiv:submit/1555588}, 2016.

\bibitem{Shawe04}
J.~Shawe-Taylor and N.~Cristianini, \emph{Kernel {M}ethods for {P}attern
  {A}nalysis}.\hskip 1em plus 0.5em minus 0.4em\relax Cambridge University
  Press, 2004.

\bibitem{reinsel98}
G.~C. Reinsel and R.~P. Velu, \emph{Multivariate {R}educed-{R}ank {R}egression:
  {T}heory and {A}pplications}.\hskip 1em plus 0.5em minus 0.4em\relax Springer
  New York, 1998.

\bibitem{Sergio15}
S.~Mu{\~n}oz-Romero, J.~Arenas-Garc{\'\i}a, and V.~G{\'o}mez-Verdejo, ``Sparse
  and kernel {OPLS} feature extraction based on eigenvalue problem solving,''
  \emph{Pattern Recognition}, vol.~48, no.~5, pp. 1797 -- 1811, 2015.

\bibitem{Roweis99}
S.~Roweis and C.~Brody, ``Linear heteroencoders,'' Gatsby Computational
  Neuroscience Unit, Tech. Rep. 1999-002, 1999.

\bibitem{Grant08}
M.~Grant and S.~Boyd, ``{CVX}: Matlab software for disciplined convex
  programming, version 2.1,'' \url{http://cvxr.com/cvx}, Mar. 2014.

\bibitem{Kim2008}
J.~Kim and H.~Park, ``Toward faster nonnegative matrix factorization: A new
  algorithm and comparisons,'' in \emph{Proc. 8th IEEE Intl. Conf. on Data
  Mining (ICDM'08)}.\hskip 1em plus 0.5em minus 0.4em\relax Pisa, Italy: IEEE,
  December 2008, pp. 353--362.

\bibitem{Pines13}
\BIBentryALTinterwordspacing
(2013) Purdue {U}niv. {W}eb site. {C}ollege of {E}ngineering. [Online].
  Available: \url{http://dynamo.ecn.purdue.edu/~biehl/MultiSpec}
\BIBentrySTDinterwordspacing

\end{thebibliography}

\bibliographystyle{IEEEtran}

\end{document}